\documentclass[10pt,twocolumn,letterpaper]{article}

\usepackage{wacv}
\makeatletter
\@namedef{ver@everyshi.sty}{}
\makeatother

\usepackage{times}
\usepackage{epsfig}
\usepackage{graphicx}
\usepackage{comment}
\usepackage{amsmath}
\usepackage{amssymb}
\usepackage{tikz}
\usetikzlibrary{positioning}
\tikzstyle{Mytext} = [text width=0.1*\linewidth]
\usepackage{cuted}
\usepackage{capt-of}

\newcommand{\gf}[2]{{\color{lightgray}{{#1}}}{\color{purple}#2}}

\def\wacvPaperID{1066} 

\wacvfinalcopy 
\wacvalgorithmstrack
\ifwacvfinal
\fi

\ifwacvfinal
\usepackage[breaklinks=true,bookmarks=false]{hyperref}
\else
\usepackage[pagebackref=true,breaklinks=true,colorlinks,bookmarks=false]{hyperref}
\fi

\ifwacvfinal
\else
\fi

\begin{document}

\title{Improving Pixel-Level Contrastive Learning by Leveraging Exogenous Depth Information} 

\author{
Ahmed Ben Saad$^{1,2}$~~Kristina Prokopetc$^{2}$~~Josselin Kherroubi$^{2}$\\ ~~Axel Davy$^{1}$~~Adrien Courtois$^{1}$~~Gabriele Facciolo$^{1}$ \vspace{0.2cm}\\
$^{1}$ENS Paris-Saclay, Centre Borelli~~~~$^{2}$Schlumberger AI Lab
}

\maketitle
\ifwacvfinal\thispagestyle{empty}\fi



\begin{strip}
\centering
\includegraphics[trim=0 0 0 12,clip,width=\linewidth]{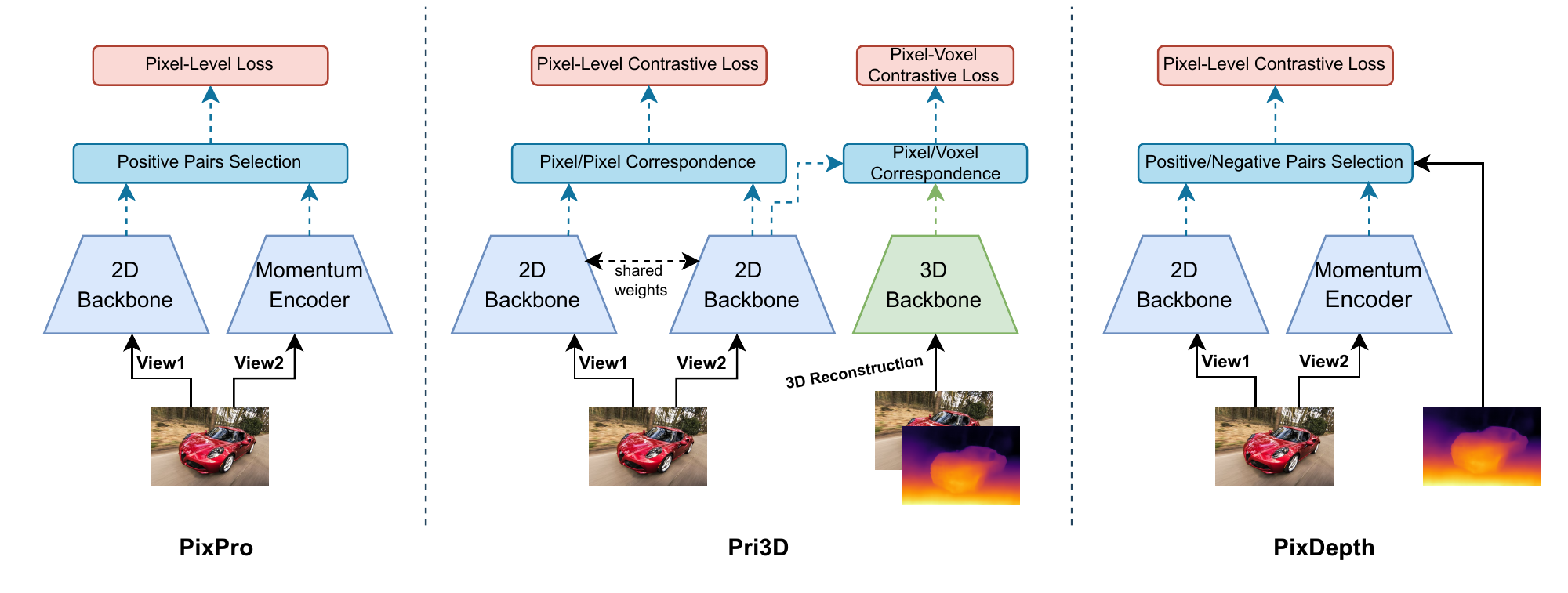}
\vspace{-2.5em}

\captionof{figure}{Visual comparison between existing methods of Contrastive Learning for Segmentation and our method: PixPro \cite{xie2021propagate} (left) does not use any 3D priors and only relies on 2D distance between image patches to distinguish between positive and negative samples. Pri3D \cite{hou2021pri3d} (middle) uses a 3D prior by reconstructing 3D scenes from stereo data and then leverages pixel to pixel and pixel to voxel correspondences. We propose to use 3D prior knowledge in a simpler way (right): A full 3D representation is not required. Instead depth maps are used in the positive/negative pairs selection, which makes it easier to train and improves results on several datasets.
The flow of data, 2D and 3D representations is represented by black arrows, blue arrows and green arrows respectively.
\label{fig:teaser}}
\end{strip}

\begin{abstract}
\vspace{-0.4cm}
Self-supervised representation learning based on Contrastive Learning (CL) has been the subject of much attention in recent years. This is due to the excellent results obtained on a variety of subsequent tasks (in particular classification), without requiring a large amount of labeled samples. 
However, most reference CL algorithms (such as SimCLR and MoCo, but also BYOL and Barlow Twins) are not adapted to pixel-level downstream tasks. 
One existing solution known as PixPro proposes a pixel-level approach that is based on filtering of pairs of positive/negative image crops of the same image using the distance between the crops in the whole image. 
We argue that this idea can be further enhanced by incorporating semantic information provided by exogenous data as an additional selection filter, which can be used (at training time) to improve the selection of the pixel-level positive/negative samples. In this paper we will focus on the depth information, which can be obtained by using a depth estimation network or measured from available data (stereovision, parallax motion, LiDAR, etc.).
Scene depth can provide meaningful cues to distinguish pixels belonging to different objects  based on their depth.
We show that using this exogenous information in the contrastive loss leads to improved results and that the learned representations better follow the shapes of objects.
In addition, we introduce a multi-scale loss that alleviates the issue of finding the training parameters adapted to different object sizes. 
We demonstrate the effectiveness of our ideas on the Breakout Segmentation on Borehole Images where we achieve an improvement of 1.9\% over PixPro and nearly 5\% over the supervised baseline. We further validate our technique on the indoor scene segmentation tasks with ScanNet and outdoor scenes with CityScapes ( 1.6\% and
\vspace{-1cm}
1.1\% improvement over PixPro respectively).
\end{abstract}

\section{Introduction}


Research in Deep Learning and Computer Vision took an important change in direction towards Unsupervised and Self-Supervised Learning \cite{wu2018unsupervised,he2020momentum,misra2020self}. This is mostly motivated by an abundance of unlabeled data in various applications and the high cost of labeling efforts. Specifically, a huge amount of unlabeled data is available from remote sensing. For example, in the petrophysical domain, the years of data acquisition with ultrasonic and other types of sensors have lead to a tremendous accumulation of data archives. Whilst the task of data labeling is not trivial, tedious and time consuming. Thus, the incentive of learning representations without the use of labels is very promising.
Following this trend, we have seen an emergence of Contrastive Learning (CL) methods~\cite{oord2018representation} as a promising set of algorithms that learns representations from data without the need for labels. These methods were adapted for Computer Vision~\cite{henaff2020data, Chen2020,caron2020unsupervised,chen2020improved} and other methods were inspired by them~\cite{grill2020bootstrap, zbontar2021barlow}, which resulted in improvements in classification tasks on ImageNet~\cite{NIPS2012_c399862d} and CIFAR-100~\cite{krizhevsky2009learning}.
The most successful CL based methods~\cite{Chen2020,he2020momentum,grill2020bootstrap} rely on two important assumptions: two transformed versions of the same image must have their representations as ``close" as possible. Similarly, different images should have their representations as ``far" as possible (even though some methods such as BYOL \cite{grill2020bootstrap} do not rely on this second assumption explicitly). 
This paradigm is called instance discrimination. According to this logic, learned representations are invariant to the set of transformations used during training. This results in instance-level features that perform well when transferred to instance-level tasks such as classification. This, however, does not encode positional information, which is crucial for pixel-level tasks~\cite{lin2017focal,liu2016ssd,ren2015faster,liu2021bootstrapping}. As a result, these representations are poorly transferable for semantic and instance segmentation. 

One proposed solution to fix this issue is PixPro \cite{xie2021propagate}.
The idea behind this method is to modify BYOL by adding a pixel-level task that uses the 2D distance between pixel-level features (belonging to two transformed image crops) to determine the positive/negative pairs. More precisely, distances smaller than a threshold are considered as positive pairs and are pulled to each other. 
They also propose an additional module 
that is used to smooth the pixel representations. While this method yields improvements on segmentation tasks compared to supervised pre-training, we believe that it can be further improved for the following reason: the positive/negative pairs selection procedure described above is equivalent to adding a shift transformation in BYOL. In fact, forcing representations of close patches to be similar makes them invariant to displacements in all directions by any value below the chosen threshold, which comes down to adding small shifts to the data augmentation. This means that we do not take into account that nearby points  might belong to different objects. For example, when some objects occlude parts of other objects. If the distance filtering threshold is not small enough or when crops are located on objects borders, this can lead to feature pairings that do not belong to the same object yet are treated as positive pairs.

\begin{figure}
    \centering
    \includegraphics[width=0.8\linewidth]{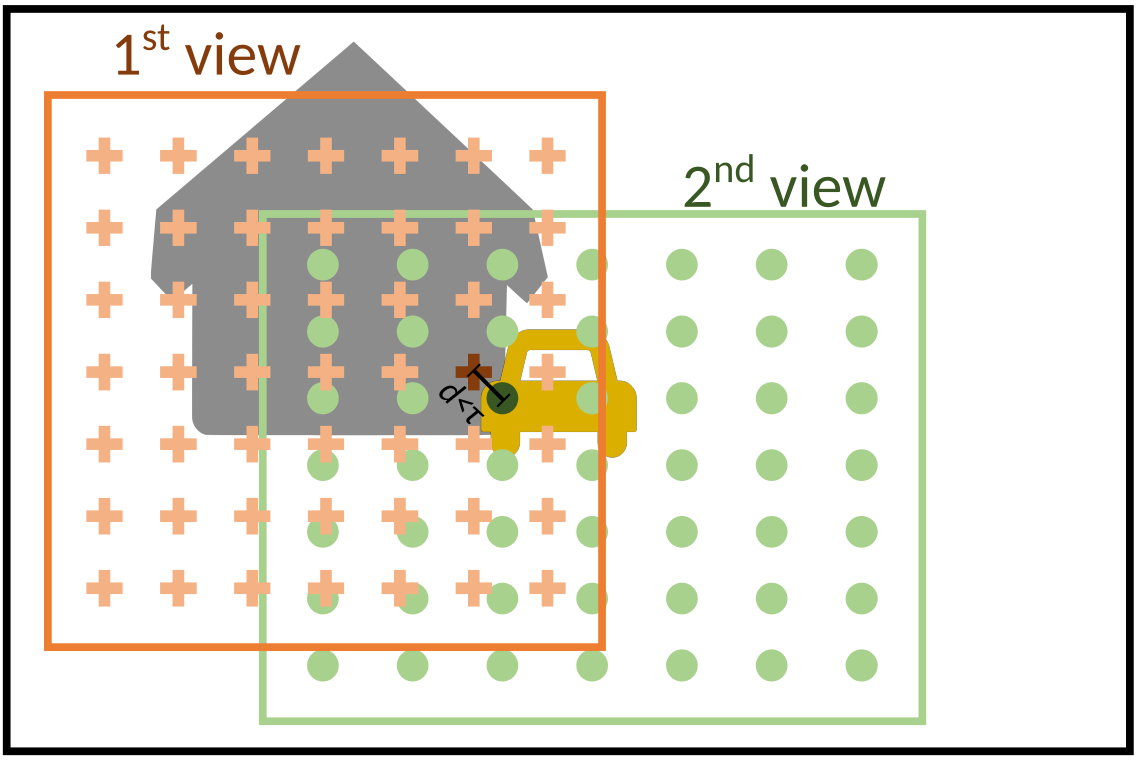}
    \caption{Example of a failure case for PixPro: the . and the + correspond to the learned representations of two views respectively. The two pixels separated by the black segment are close on the image ($d<\tau$) despite belonging to two different objects. This figure is based on Figure~1 in~\cite{xie2021propagate}}
    \label{fig:problempixelcl}
\end{figure}
To alleviate this problem, we propose PixDepth. The main idea is to leverage exogenous depth information in the positive/negative pairs selection. The image depth does not only provide very useful prior semantic information about the correct object boundaries, but also prevents learning fuzzy representations near the object boundaries.
Specifically, when comparing pixel-level features, in addition to the 2d distance we also compute the difference in depth at the corresponding locations. If the depth difference is larger than a depth threshold we assume that the objects are distinct. This additional constraint refines the choice of positive pairs, thus injecting more semantic information in the training. We demonstrate our approach by pre-training on the ImageNet dataset \cite{deng2009imagenet} and using depth maps pre-computed using MIDAS~\cite{ranftl2020towards} and then fine-tuning on the CityScapes dataset \cite{cordts2016cityscapes}. Moreover, we show that we can drop the pre-trained monocular depth estimator and directly use  the large amount of existing binocular stereo data to pre-train PixDepth. In addition, we will discuss how the threshold parameters may depend on the scale of the objects in the dataset and propose a multi-threshold loss that reduces this dependency.

\begin{figure*}
    \centering
    \includegraphics[width=\linewidth,height=6cm]{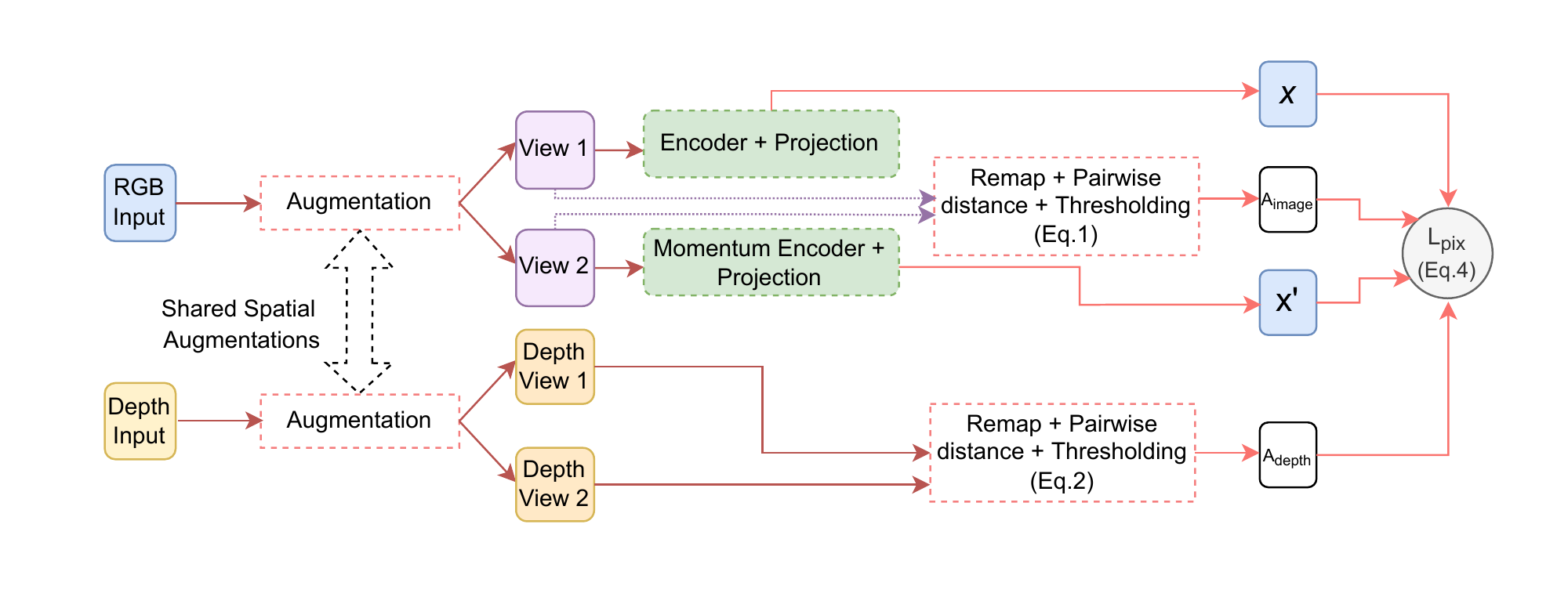}
    \vspace{-3em}
    \caption{The detailed overview of our proposed framework (PixDepth): given an input image, we apply a set of augmentations and then crop two patches (as in \cite{xie2021propagate}). We do the same thing for the depths views. Next, each view goes through the target or the momentum network. To build the positive and negative pairs, the depth views are remapped to the feature map sizes and then a pairwise absolute difference (Equation \ref{eq:depthresh}) is calculated to measure the distance in depth space between image patches. This newly obtained mask $\mathcal{A}_{depth}$ helps the network to better distinguish if the selected patches are of the same object or not.}
    \label{fig:overview}
\end{figure*}

We demonstrate the application of our ideas with experiments on natural images from ImageNet~\cite{deng2009imagenet} and ScanNet~\cite{dai2017scannet} (using monocular depth estimations and measured depth data for pretraining respectively) as well as on an industrial use case with borehole images. In this case we will use the transit time of ultrasonic waves as an indicator of depth. We also present an experimental setup to qualitatively analyze the representations learned by PixPro and PixDepth.
This work is motivated by the geological features detection in petrophysical domain and can be easily transferred to other applications where this external information is widely available in contrast to labeled data.
%
%
To summarize, our contributions are:
\vspace{-.5em}\begin{itemize}
\setlength\itemsep{-.3em}
    \item Improving PixPro by making use of the prior knowledge of depth to improve the learned representations.
    \item A   multi-scale loss and training with multiple distance and depth thresholds that helps encoding objects with different scales.
    \item An extensive experimental evaluation to support our findings on different datasets for semantic segmentation on CityScapes and ScanNet and geological features identification on a BoreholeImage dataset.

\end{itemize}

\section{Related Work}
\paragraph{Contrastive Learning and Computer Vision}
There exist many works which apply CL to visual tasks \cite{henaff2020data, Chen2020,caron2020unsupervised,chen2020improved,he2020momentum}. Most notably SimCLR  \cite{Chen2020} proposes a simple CL framework that surpassed the supervised pre-training in ImageNet classification but required a very large batch size to ensure good performances. BYOL \cite{grill2020bootstrap}, on the other hand, is not an explicit CL method but was inspired from SimCLR and uses a momentum encoder to remove the need of large batch sizes. The main limitation of these methods is inability to generate instance-level features that transfer well for pixel-level tasks. When it comes to image segmentation, \cite{xie2021propagate} proposes a modification to BYOL by adding a pixel-level pretext task that separates positive and negative pairs of two crops in one image by computing the distance between these crops. If this distance is smaller than a given threshold, the pair of crops is considered as positive example and a negative example otherwise. Our method extends this idea in two ways: we improve the  selection of pairs by leveraging depth information and we incorporate different thresholds at the same time to remove the dependence on thresholds as hyperparameters by taking multiple scales into account.

Pri3D~\cite{hou2021pri3d} is based on a similar intuition of using prior 3D information to improve CL. The authors rely on a 3D reconstruction of indoor scenes to build another proposed pixel-level pretext task. This creates pixel-to-pixel correspondences from different views of the same scene and pixel-to-voxel correspondences between representations of 2D views and 3D meshes. While this idea is comparable to ours, we believe that our framework is simpler because we do not rely on a 3D reconstruction and a neural network to encode this 3D structure. Moreover, depth maps are more readily available (from RGB-D sensors, stereovision, monocular estimation) than the full 3D scenes used by Pri3D.

Recently, Point-level Region Contrast~\cite{bai2022point} has proposed a CL approach at region level by directly sampling individual point pairs
from different regions that are defined as a fixed grid at the beginning of the training and are updated periodically during training. In contrast, we believe that the depth information provides a strong semantic hint for segmenting objects as they have distinct depths from their background.

\paragraph{Depth Estimation}
Visual ability to perceive the world in 3D and the ability to estimate the distance/depth of an object from the source is paramount in numerous tasks including scene understanding. The details in the environment that allow to perceive depth are often called depth cues. They can be categorised as \emph{binocular} cues, when viewing a scene with both eyes, as \emph{motion} cues, resulting from the motion of the observer or in the scene, or as \emph{monocular} cues, when viewing a static scene with one eye. Binocular stereo-based depth estimation is a go-to solution when more than one observation of the scene is available. To achieve this, matching pixels from the left and right camera images are identified and the disparity (or difference) in corresponding pixel locations can then be used to infer the depth component. However, in many application scenarios only one observation of a scene is available at each time. To this end, monocular depth estimation methods such as MRF based formulations \cite{saxena2008make3d}, methods based on geometry assumptions \cite{hoiem2005automatic,poggi2018learning} or non-parametric methods \cite{karsch2014depth,xu2018non}.
More recent deep learning based techniques \cite{ranftl2020towards,miangoleh2021boosting} showcase the great progress in ability to estimate disparities and depth maps from a single image. In our work we use MIDAS \cite{ranftl2020towards} a recently proposed method based on a transformer architecture, which was trained on a very large selection of diverse datasets.

\section{Contrastive Learning with PixDepth}


Our method builds upon the pixel contrast loss introduced in~\cite{xie2021propagate}. This loss is defined on pixel-features extracted from two augmented views of the same image. 
More precisely, given an input image, 
two color-augmented crops are generated. The two crops are further resized to a fixed resolution and passed through a regular encoder network and a momentum encoder network~\cite{he2020momentum,grill2020bootstrap}. The obtained feature maps (\textit{e.g.} of size $1024 \times 8 \times 8$) are then projected using a pixel level projector (two $1\times 1$ convolution layers). The output of the pixel projector already contains 2D positional information. 

Figure~\ref{fig:problempixelcl} illustrates the features for two crops. The plus signs and circles correspond to the vectors of the first and second representation vectors of the respective first and second views. 
The idea of~\cite{xie2021propagate} is to use all the pairs of pixel-features from the two crops to generate the positive and negative pairs for contrastive learning. Intuitively, overlapping or spatially close locations are used for positive pairs. This is achieved by constructing a positive/negative mask
\begin{equation}
    A_{image}(i,j) = \begin{cases} 1, & \mbox{if } dist(i,j) \leq \mathcal{T} \\ 0, & \mbox{if } dist(i,j) > \mathcal{T} \end{cases},
    \label{eq:pixpro_threshold}
\end{equation}
where $\mathcal{T}$ is the distance threshold, $dist$ is the normalized (from 0 to 1) euclidean distance between 2D point coordinates, $i$ and $j$ are the indexes of vectors in the first and second view respectively. 

The pixel contrast loss functions~\cite{xie2021propagate} is then defined as
\begin{equation}
\label{eqn:cosine-softmax}
\small
    \mathcal{L}_\text{Pix}(i) =  - {\log \frac{\sum \limits_{j\in \Omega_p^i} e^ { { \cos \left( {\mathbf{x}_{i}},{\mathbf{x}'_{j}} \right)} /{\tau} }}{\sum \limits_{j\in \Omega_p^i} e^{ { \cos \left( {\mathbf{x}_{i}},{\mathbf{x}'_{j}} \right)}/ {\tau}}  + \sum\limits_{k\in \Omega_n^i}e^{ { \cos \left( {\mathbf{x}_{i}},{\mathbf{x}'_{k}} \right)}/ {\tau}}}},
\end{equation} 
where $i$ is a first-view pixel that is also present in the second view; $\Omega_p^i$ and $\Omega_n^i$ are groups of pixels in the second view that have been designated as being positive (when $A_{image}(i,j)=1$) and negative (when $A_{image}(i,j)=0$), respectively, in relation to pixel $i$. The pixel feature vectors in two views are $\mathbf{x}_i$ and $\mathbf{x}_j'$, and $\tau$ is a scalar temperature hyper-parameter. All pixels from the first view that are located at the intersection of the two views are averaged to determine the loss. Similar to the first view, the second view computes and averages the contrastive loss for each pixel~$j$. The average of all image pairs in a mini-batch represents the final loss. The overall framework is illustrated in the upper half of Figure~\ref{fig:overview}.

Note that this contrastive loss forces feature representations corresponding to the same spatial location (or close depending on the choice of $\mathcal{T}$) to be close.





We believe, however, that this training process has two flaws: 
1) The authors use a single pixel distance threshold $\mathcal{T}$  in the pairs selection which is a dataset-specific hyperparameter and is based on a strong assumption that nearby pixels belong to the same object; 
2) This logic does not reflect a structural similarity/difference in cropped patches, which may potentially produce contradictory signals, especially on the frontier between objects in one image.
The second problem is particularly important and is illustrated in Figure~\ref{fig:problempixelcl}. If we look at the vectors distinguished by dark colors, we notice that they belong to two different objects, despite their proximity in the image. PixPro will tend to consider this pair of vectors as positive and this is detrimental to the quality of the representations.

To resolve this, we propose to modify the procedure for assigning pairs of positive or negative vectors so that it can incorporate some additional semantic information. Depth maps are a good solution because changes in depth can be an indicator of object change. We have therefore considered depth maps as an additional assignment filter. 

\subsection{Incorporating Depth Information}

Figure~\ref{fig:overview} shows a general overview of our method.
We enhance~\cite{xie2021propagate} by taking crops from the depth maps at the same positions of our two views. After that, we map them to the size of the feature maps (\textit{e.g.} $7\times 7$) and we calculate the pairwise difference between the pixels of the resized depth crops.
In order to prevent crops that are distant in 2D and have similar depth (\textit{e.g.} two different objects in foreground). We apply this new selection on top of the previous one. Thus, we build a new positive/negative depth mask as: 
\begin{equation}
    A_{depth}(i,j) =  \, \begin{cases} 1, & \mbox{if } |depth(i) - depth(j)| \leq \mathcal{T'} \\ 0, & \mbox{if } |depth(i) - depth(j)| > \mathcal{T'} \end{cases},
    \label{eq:depthresh}
\end{equation}
where $\mathcal{T'}$ is the threshold on the depth maps
and  the final selection mask is then obtained by element-wise multiplication of both masks
\begin{align}
 A_{final}(i,j) = & \, A_{image}(i,j)\times A_{depth}(i,j).
\end{align}

The mask $A_{final}$ is then used to define the groups of positive/negative pixels $\Omega_p^i$ and $\Omega_n^i$ to be used in the loss~\eqref{eqn:cosine-softmax}.

In summary, this is a way of incorporating depth prior information into the learning procedure without the need for the encoder to ``see" the depth maps at any level of the training. It means that no extra computing power nor time is needed. Simultaneously, the selection of positive and negative pairs is now based on semantic information given by these maps and not only a distance in a projected 2D image, which might end up mixing different objects. As we will see in the next section, the depth information can be found in different sources such as depth maps, disparity maps, etc. Thus, we chose to normalize their values between 0 and 1. It is important to note that we are only interested in differences between depth values and not the actual depth measurements/estimations. Consequently, converting disparity maps to depth maps, for example, is not required.

\subsection{Multi-thresholds for Different Scales}

\begin{figure}
    \centering
    \includegraphics[trim=20 60 20 0,clip,width=\linewidth]{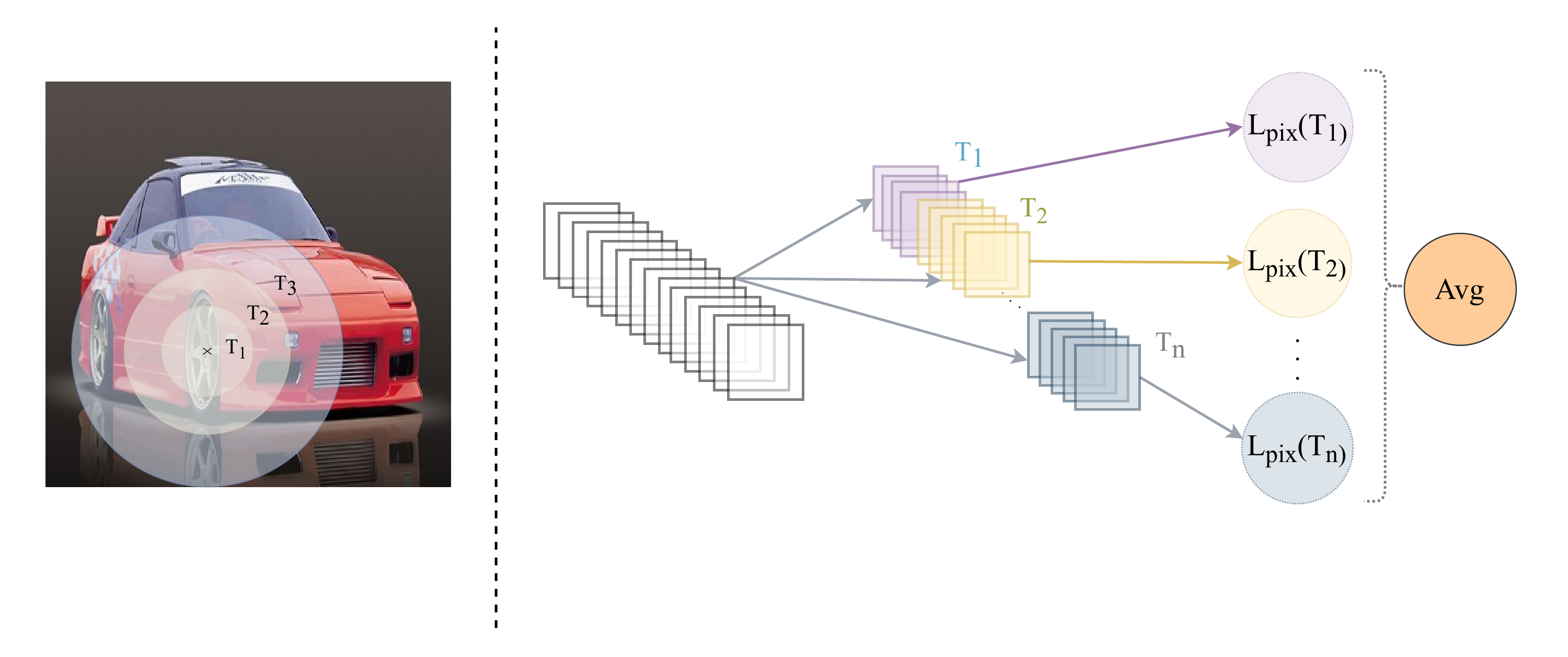}
    \caption{An overview of the multi-thresholding setup to account for different scale. We want our representations to be aware of the relative context at different scales. In the left figure, we show how each threshold $\bigl\{\mathcal{T}_i\bigr\}_{1 \leq i \leq n} $ (we arbitrarily chose $n=3$ and $\mathcal{T}_1 \leq \mathcal{T}_2 \leq \mathcal{T}_3$ for visualization purposes) affects the pairs selection procedure: If we take a crop centered at the X mark, each disk represents the centers of the crops that will form a positive pair with the reference crop. In practice (right) , we divide the output of our encoder and projector channel-wise into $n$ equal sized feature maps and then we independently apply Eq.\ref{eq:pixpro_threshold} on each one using its corresponding threshold. We than average all the contributions for the final loss computations.}
    \label{fig:multithresh}
\end{figure}

The process of pairs selection described above depends on two hyperparameters $\mathcal{T}$ and $\mathcal{T'}$. The choice of these two parameters is tricky and may negatively affect the quality of the learned representations. In fact, objects come in different shapes and sizes depending on datasets and may vary greatly within a single dataset. This leads to the optimal values of $\mathcal{T}$ and $\mathcal{T'}$ being dependent on the pre-training dataset (see Table~\ref{tab:threshdataset}). To alleviate this problem without the need of extra processing, we propose the following procedure (illustrated in Figure~\ref{fig:multithresh}) that allows different parts of the feature representations to be discriminating at different scales.

\begin{table}
\centering
\begin{tabular}{ccc}
\hline
Pre-traning dataset & $\mathcal{T'}$ & mIoU \\ \hline
Imagenet  & 0.3 & 74.3 \\
+ Depth with MIDAS & 0.5 & \textbf{76.9} \\
 & 0.7 & 73.9 \\ \hline
BoreholeImage & 0.3 & \textbf{74.0} \\
 & 0.5 & 72.8 \\
 & 0.7 & 70.7 \\ \hline
\end{tabular}
\caption{The effect of varying $\mathcal{T'}$ on the fine-tuning performances. The fine-tuning was done on CityScapes for the first 3 rows and The labeled BoreholeImage dataset for the rest.}
\label{tab:threshdataset}
\end{table}

Given an output feature map $\mathcal{F}$ of size $C \times H \times W$, we divide it into $n$ disjoint features maps $\bigl\{\mathcal{F}_i\bigr\}_{1 \leq i \leq n} $ of size $\frac{C}{n}\times H \times W$. For each $\mathcal{F}_i$ we apply a different threshold $\bigl\{\mathcal{T}_i\bigr\}_{1 \leq i \leq n} $ and $\bigl\{\mathcal{T'}_i\bigr\}_{1 \leq i \leq n} $. Every $\mathcal{F}_i$ is then treated as an independent feature vector. This leads to computing $n$ loss terms as in Equation~\ref{eqn:cosine-softmax}:
\begin{equation}
\resizebox{0.9\hsize}{!}{$
    \mathcal{L}_\text{Pix}^k(i)= -{\log \frac{\sum \limits_{j\in \Omega_p^i} e^ { { \cos \left( {\mathbf{x^k}_{i}},{\mathbf{x^k}'_{j}} \right)} /{\tau} }}{\sum \limits_{j\in \Omega_p^{i,k}} e^{ { \cos \left( {\mathbf{x^k}_{i}},{\mathbf{x^k}'_{j}} \right)}/ {\tau}}  + \sum\limits_{m\in \Omega_n^{i,k}}e^{ { \cos \left( {\mathbf{x^k}_{i}},{\mathbf{x^k}'_{m}} \right)}/ {\tau}}}},
    $}
\end{equation}
where $x_{i/j/m}^k$ represents the $k$\textsuperscript{th} portion of the feature vector $x_{i/j/m}^k$ (the portion of $x$ that lies in $\mathcal{F}_k$). The groups of pixels $\Omega_{n/p}^{i,k}$ are obtained differently for each $k$ using the threshold $\mathcal{T}_k$. The remaining parameters are the same as in Equation~\ref{eqn:cosine-softmax}
and the total loss is the average of all these contributions
\begin{equation}
\small
    \mathcal{L}_\text{PixMulti}(i)  =  \frac{1}{n}\sum_{k=1}^n \mathcal{L}_\text{Pix}^k(i).
\end{equation}

In summary, each part of our new feature vector will encode a different scale centered around the object of interest. By selecting thresholds at different scales, we remove the dependency of our method on this hyperparameter. In practice, we  choose 3 values for ${\mathcal{T}_i}$ and ${\mathcal{T'}_i}$:  $0.3$, $0.5$, $0.7$. the values are chosen between 0 and 1 because both the depth maps and the 2D distance defined in \cite{xie2021propagate} are normalized and thus all distances are in the same interval. We believe that choosing 3 values that account for 3 different scales on the image is sufficient for multi-thresholding (see next Section).
When we use this method on both hyperparameters in the same time, we use 
$\bigl\{\mathcal{T}_i, \mathcal{T'}_i\bigr\}_{1 \leq i \leq n} $ for each $\mathcal{F_i}$. This limits the possible combinations of threshold pairs (and thus the feature map division).

\section{Experiments}
The main motivation for this work is driven by the geological feature segmentation in remote sensing and more specifically petrophysical domain, which prompts a set of experiments on ultrasonic borehole image dataset that we have collected on the course of this study. This is complemented by a set of experiments on datasets from other applications such as semantic segmentation on CityScapes and ScanNet, which helps us validate the generality of our contributions and facilitates the reproducibility of our results. In the following sections we explain the experimental setup and describe in detail the datasets used in our study, configuration of the training set up as well as our approach for visually inspecting the quality of learned representations.  
\begin{figure}
    \centering
    \includegraphics[trim=0 50 0 50,clip, width=\linewidth, height=2.25cm]{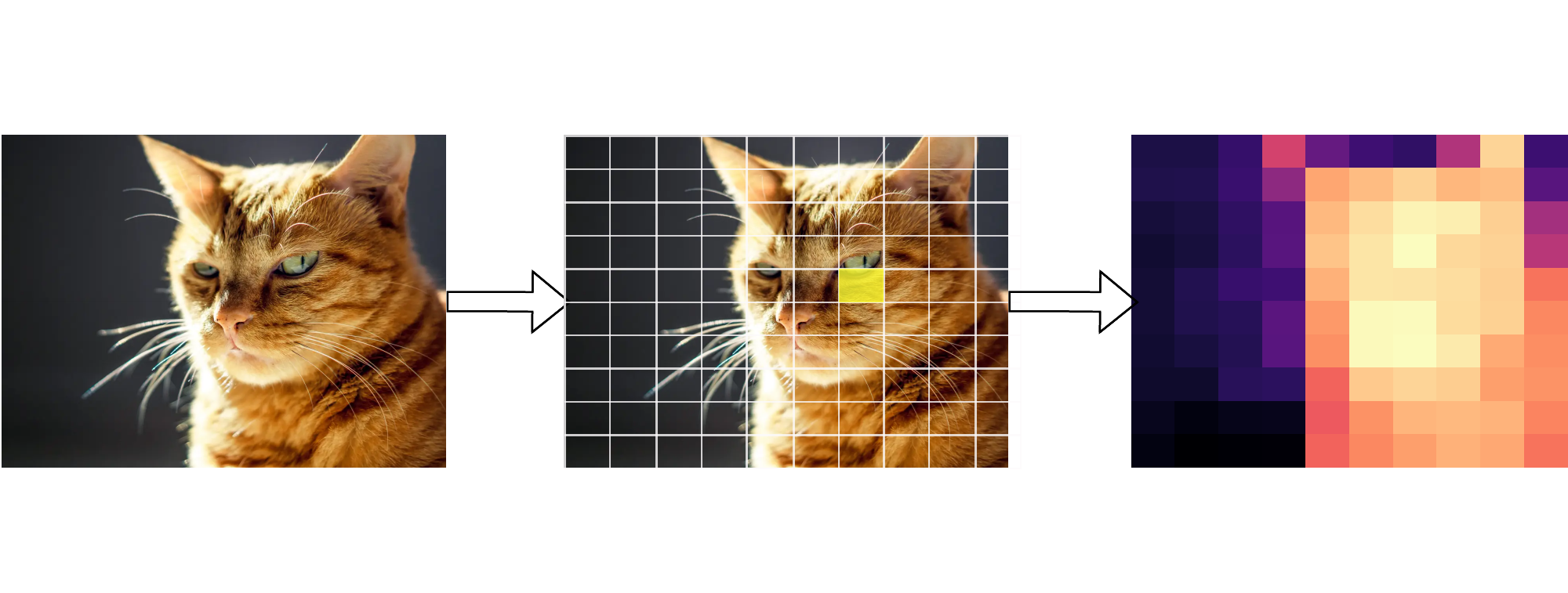}

    \caption{An overview of the representation quality check: an input image (left) is fed into the encoder to produce a feature map (middle dots). Each vector of this feature approximately encodes a patch of the image (a cell of the grid). We select one of the vectors (in yellow) and we measure its similarity to the rest of the vectors. We then plot a similarity heatmap (right).  }
    \label{fig:qlttveexp}
\vspace{-0.3cm}
\end{figure}

\subsection{Data}
\begin{figure}[]
    \centering

    \begin{tikzpicture}[
 image/.style = {text width=0.11\textwidth, 
                 inner sep=0pt, outer sep=0pt},
node distance = 1mm and 1mm
                        ] 
\node [image] (frame1)
    {\includegraphics[height=\linewidth,width=\linewidth]{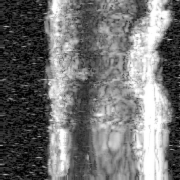}};
\node [image,right=of frame1] (frame2) 
    {\includegraphics[height=\linewidth,width=\linewidth]{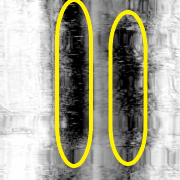}};
\node [image,right=of frame2] (frame3) 
    {\includegraphics[height=\linewidth,width=\linewidth]{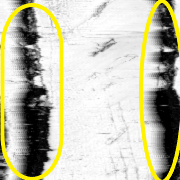}};
\node [image,right=of frame3] (frame4) 
    {\includegraphics[height=\linewidth,width=\linewidth]{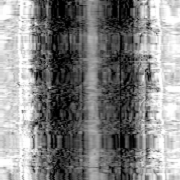}};

\node[image,below=of frame1] (frame5)
    {\includegraphics[height=\linewidth,width=\linewidth]{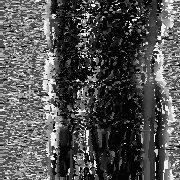}};
\node[image,right=of frame5] (frame6)
    {\includegraphics[height=\linewidth,width=\linewidth]{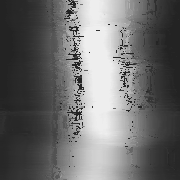}};
\node[image,right=of frame6] (frame7)
    {\includegraphics[height=\linewidth,width=\linewidth]{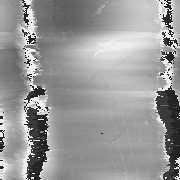}};
\node[image,right=of frame7] (frame8)
    {\includegraphics[height=\linewidth,width=\linewidth]{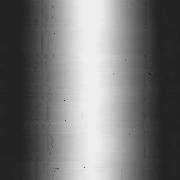}};

\end{tikzpicture}
    \caption{Examples of borehole images from UK NDR dataset used in our studies. First row shows examples of the images based on measured amplitudes of ultrasonic waves. Second row shows corresponding images based on transit times estimation. Breakouts identified on amplitude images are highlighted in yellow.}
    \label{fig:dsexamples}
\end{figure}

\noindent\textbf{BoreholeImage} dataset used in our study is a basis for geo-mechanical and geological interpretation of hydrocarbon reservoirs based on ultrasonic (US) imaging and plays an important role in Oil and Gas industry context. Borehole US images show meaningful information about petrophysical properties such as breakouts - a type of stress-induced geological features. Thus, the dataset consists of ``unwrapped” measures of the amplitude of an ultrasonic wave sent towards a geological formation and received back by an ultrasonic sensor which is then represented in 2D format. This dataset is partially based on images obtained from a public repository UK NDR\footnote{\url{https://ndr.ogauthority.co.uk}} and on data provided by industrial operators. Figure~\ref{fig:dsexamples} demonstrates several examples of borehole images that correspond to the UK NDR portion of the dataset. We have used a diverse selection of nearly 27k images collected from 140 different wells. This UK NDR portion of the dataset does not provide labels whilst pixel-level annotations for the breakouts are coming from the remaining portion. 

\noindent\textbf{CityScapes} is a large-scale dataset that contains a diverse set of binocular stereo-vision sequences recorded in street scenes from 50 different cities, with high quality pixel-level annotations of 5000 frames in addition to a larger set of 20000 weakly annotated frames~\cite{cordts2015cityscapes}.

\noindent\textbf{ScanNet} is an RGB-D video dataset containing 2.5 million views in more than 1500 scans, annotated with 3D camera poses, surface reconstructions, and instance-level semantic segmentations. \cite{dai2017scannet}.

\subsection{Instance and Semantic Segmentation}

We start by pre-training a ResNet \cite{DBLP:journals/corr/HeZRS15} encoder on one of the unlabeled datasets depending on the subsequent task (i.e. indoor scene segmentation on CityScapes and ScanNet or breakouts identification on BoreholeImage dataset). Then, we integrate this pre-trained encoder into a semantic segmentation architecture where we freeze its weights. In this work we employ DeepLabV3~\cite{chen2017rethinking} as a segmentation model. Next, we proceed by training the segmentation model on the corresponding labeled portion of the previously used dataset. To demonstrate the improvements due to our approach and for fair comparison we report quantitative results for fully supervised trainings, compared with the original PixPro and our PixDepth pre-trainings.

For the Breakouts segmentation on BoreholeImage dataset we used a ResNet-34 encoder and two splits of the dataset for pre-training and fine-tuning. The results are shown in Table~\ref{tab:results2}. For the  experiments on the CityScapes data we also used a ResNet-34 encoder but we perform two variations of the experiment. In the first variation we used ImageNet-1K as a pre-training dataset with monocular depths maps computed using MIDAS~\cite{ranftl2020towards}. In the second variation we remove the dependency on a pre-trained monocular depth estimation network (MIDAS) and instead  directly use a subset of the disparity maps that were used to train it. The corresponding results are shown in Table~\ref{tab:results1}. For the experiments on the ScanNet dataset we used a ResNet-18 encoder and the same dataset for pre-training and fine-tuning. The results are shown in Table~\ref{tab:results3}.

\begin{table}
\resizebox{\columnwidth}{!}{%
\begin{tabular}{llllll}
\hline
Pre-training Dataset & Pre-training Method  & IoU \\ \hline
Imagenet & supervised & 69.1 \\
BoreholeImage  & PixPro &  72.1 \\
BoreholeImage  + Transit Time & PixDepth & \textbf{74.0} \\ \hline
\end{tabular}%
}
\caption{Breakout segmentation task with BoreholeImage dataset}
\label{tab:results2}
\vspace{-.5em}
\end{table}

\begin{table}
\resizebox{\columnwidth}{!}{%
\begin{tabular}{lllll}
\hline
Pre-training Dataset & Pre-training Method & mIoU \\ \hline
Imagenet & supervised   & 71.6 \\
Imagenet & PixPro & 75.8 \\
Imagenet + depths maps & PixDepth   & \textbf{76.9} \\
35\% of Midas training set & PixDepth  & \underline{76.3} \\ \hline
\end{tabular}%
}
\caption{Indoor segmentation task with CityScapes dataset. }
\label{tab:results1}
\vspace{-1em}
\end{table}

\begin{table}
\centering
\begin{tabular}{ll}
\hline
 Pre-training Method  & IoU  \\ \hline
Supervised & 44.7 \\ 
PixPro    & 46.3 \\ 
Pri3D    & 48.1 \\ 
PixDepth  & \textbf{49.7} \\ \hline
\end{tabular}
\caption{Indoor segmentation task with ScanNet dataset.}
\label{tab:results3}
\vspace{-0.5cm}
\end{table}

\begin{figure*}
    \centering
    \resizebox{\linewidth}{!}{
    \begin{tikzpicture}[
 image/.style = {text width=0.1\textwidth,  inner sep=0pt, outer sep=0pt},
node distance = 1mm and 1mm
                  ] 
\node [image] (frame1)
    {\includegraphics[width=\linewidth]{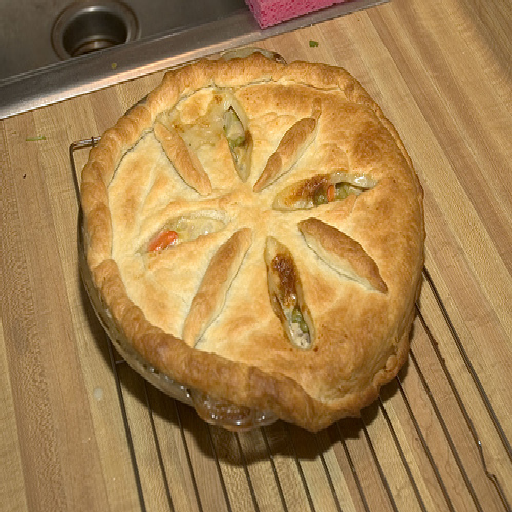}};
    
\node [image,right=of frame1] (frame2) 
    {\includegraphics[width=\linewidth,height=\linewidth]{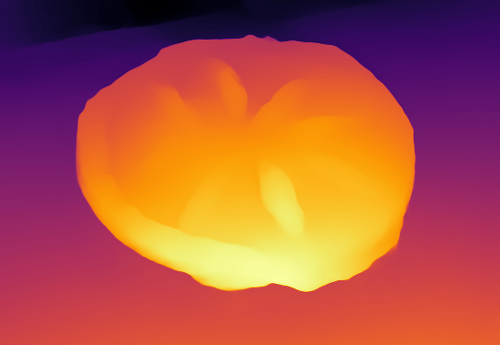}};
\node [image,right=of frame2] (frame222) 
    {\includegraphics[width=\linewidth]{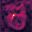}};
\node [image,right=of frame222] (frame3) 
    {\includegraphics[width=\linewidth]{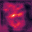}};
\node[image,below=of frame1] (frame4)
    {\includegraphics[width=\linewidth]{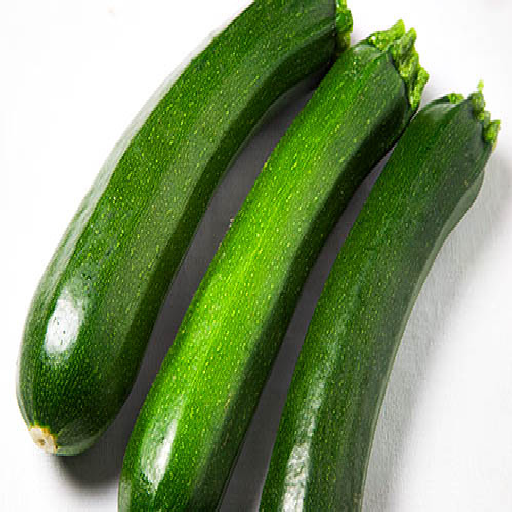}};
\node[image,right=of frame4] (frame44)
    {\includegraphics[width=\linewidth,height=\linewidth]{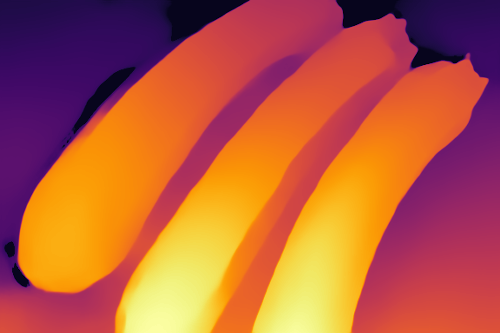}};
\node[image,right=of frame44] (frame5)
    {\includegraphics[width=\linewidth]{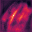}};
\node[image,right=of frame5] (frame6)
    {\includegraphics[width=\linewidth]{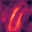}};
\node[image,below=of frame4] (frame7)
    {\includegraphics[width=\linewidth]{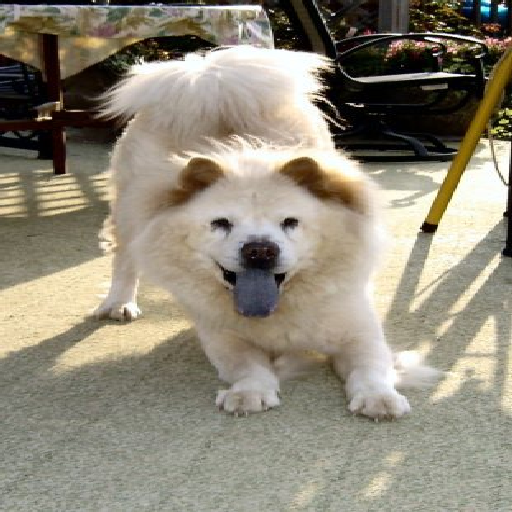}};
\node[image,right=of frame7] (frame77)
    {\includegraphics[width=\linewidth, height=\linewidth]{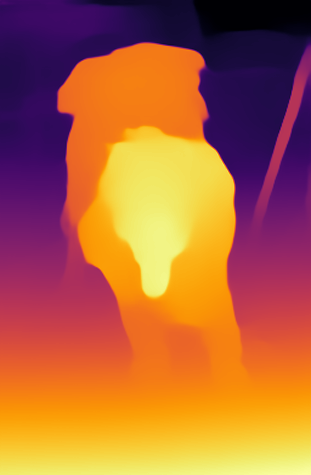}};
\node[image,right=of frame77] (frame8)
    {\includegraphics[width=\linewidth]{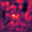}};
\node[image,right=of frame8] (frame9)
    {\includegraphics[width=\linewidth]{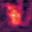}};

\node[image,below=of frame7] (frame10)
    {\includegraphics[width=\linewidth]{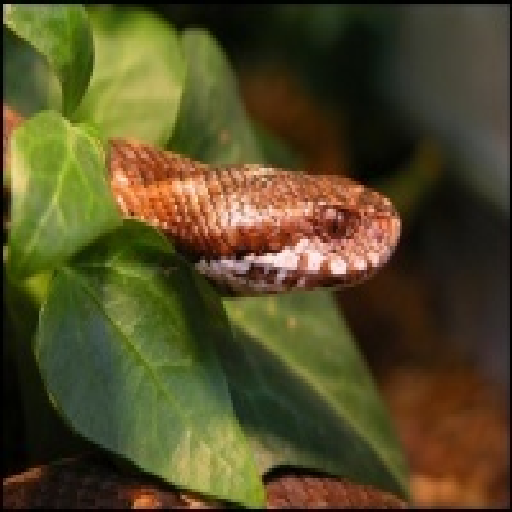}};
\node[image,right=of frame10] (frame11)
    {\includegraphics[width=\linewidth, height=\linewidth]{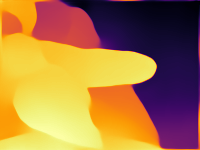}};
\node[image,right=of frame11] (frame12)
    {\includegraphics[width=\linewidth]{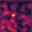}};
\node[image,right=of frame12] (frame13)
    {\includegraphics[width=\linewidth]{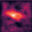}};

\node[image,right=0.2cm of frame3] (frame14)
    {\includegraphics[width=\linewidth]{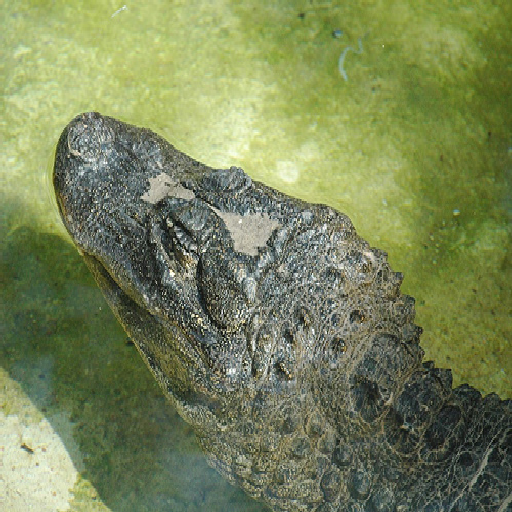}};
\node[image,right=of frame14] (frame15)
    {\includegraphics[width=\linewidth, height=\linewidth]{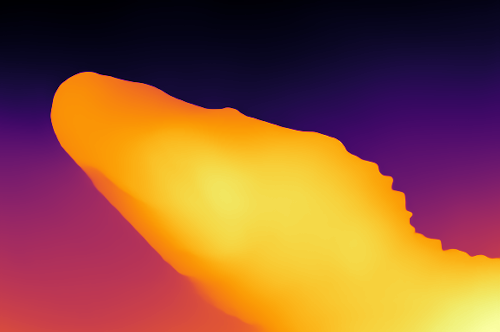}};
\node[image,right=of frame15] (frame16)
    {\includegraphics[width=\linewidth]{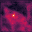}};
\node[image,right=of frame16] (frame17)
    {\includegraphics[width=\linewidth]{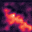}};

\node[image,below=of frame14] (frame18)
    {\includegraphics[width=\linewidth]{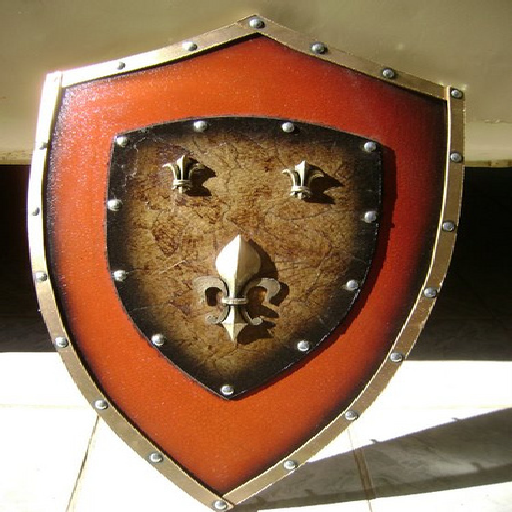}};
\node[image,right=of frame18] (frame19)
    {\includegraphics[width=\linewidth, height=\linewidth]{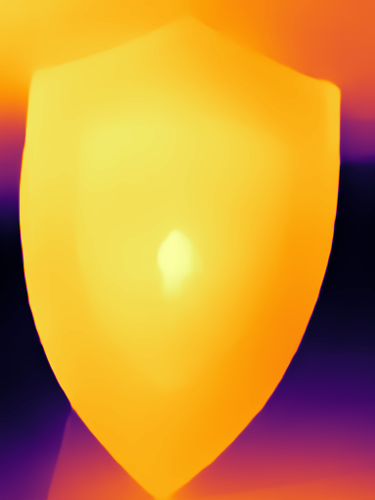}};
\node[image,right=of frame19] (frame20)
    {\includegraphics[width=\linewidth]{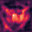}};
\node[image,right=of frame20] (frame21)
    {\includegraphics[width=\linewidth]{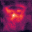}};

\node[image,below=of frame18] (frame22)
    {\includegraphics[width=\linewidth]{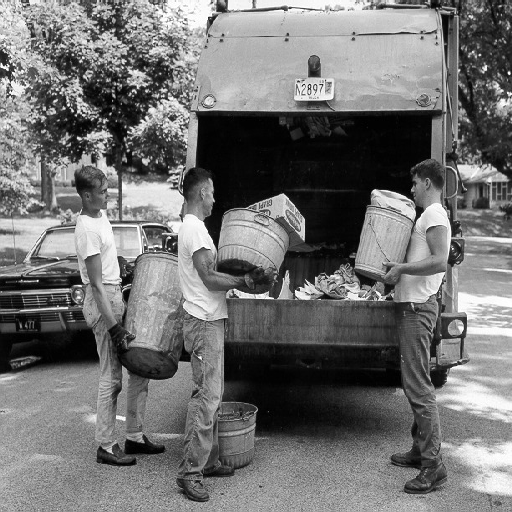}};
\node[image,right=of frame22] (frame23)
    {\includegraphics[width=\linewidth, height=\linewidth]{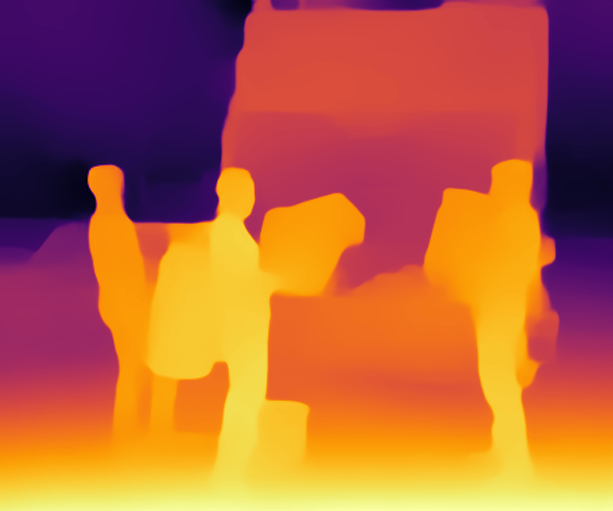}};
\node[image,right=of frame23] (frame24)
    {\includegraphics[width=\linewidth]{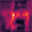}};
\node[image,right=of frame24] (frame25)
    {\includegraphics[width=\linewidth]{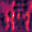}};

\node[image,below=of frame22] (frame26)
    {\includegraphics[width=\linewidth]{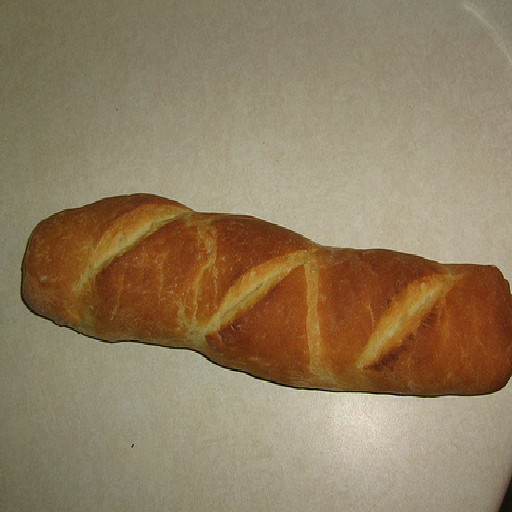}};
\node[image,right=of frame26] (frame27)
    {\includegraphics[width=\linewidth, height=\linewidth]{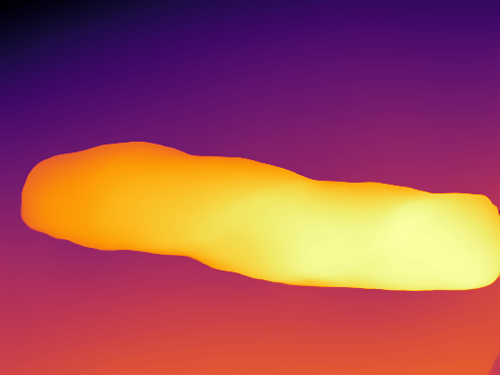}};
\node[image,right=of frame27] (frame28)
    {\includegraphics[width=\linewidth]{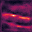}};
\node[image,right=of frame28] (frame29)
    {\includegraphics[width=\linewidth]{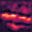}};

\node[image,below=of frame26] (frame30)
    {\includegraphics[width=\linewidth]{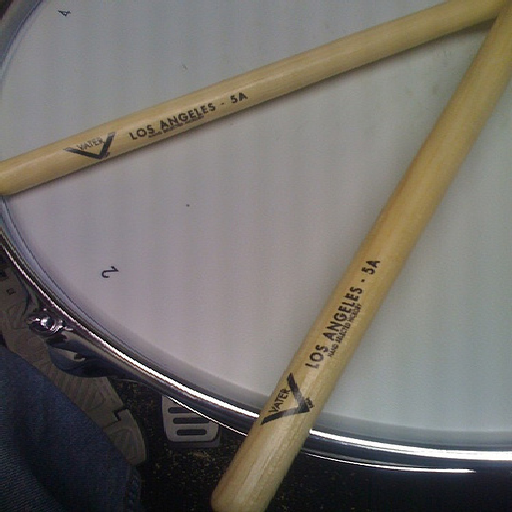}};
\node[image,right=of frame30] (frame31)
    {\includegraphics[width=\linewidth, height=\linewidth]{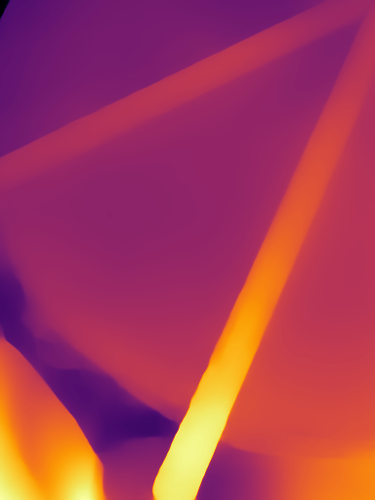}};
\node[image,right=of frame31] (frame32)
    {\includegraphics[width=\linewidth]{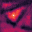}};
\node[image,right=of frame32] (frame33)
    {\includegraphics[width=\linewidth]{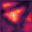}};

\node[image,below=of frame10] (frame34)
    {\includegraphics[width=\linewidth]{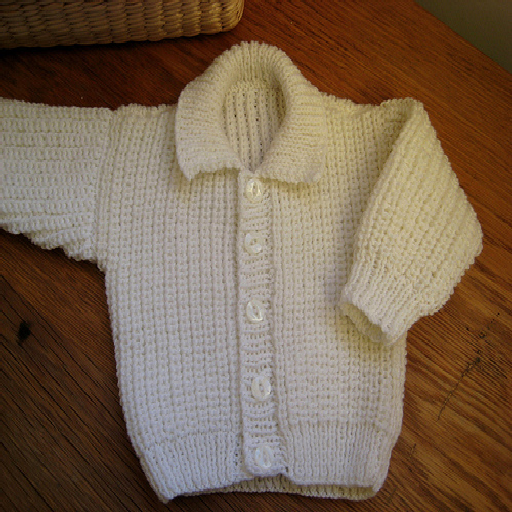}};
\node[image,right=of frame34] (frame35)
    {\includegraphics[width=\linewidth, height=\linewidth]{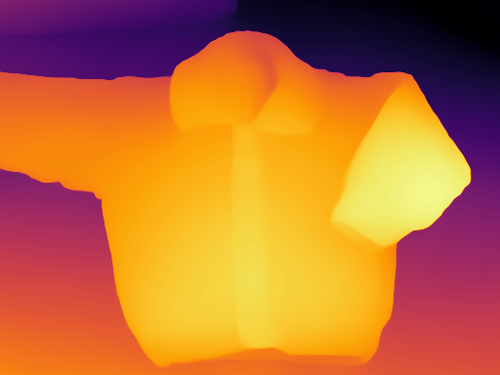}};
\node[image,right=of frame35] (frame36)
    {\includegraphics[width=\linewidth]{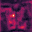}};
\node[image,right=of frame36] (frame37)
    {\includegraphics[width=\linewidth]{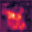}};

\node[Mytext,above= of frame1]  {\scriptsize \vphantom{p}Input Image\vphantom{p}};
\node[Mytext,above= of frame222]  {\scriptsize \vphantom{p}PixPro\vphantom{p}};

\node[Mytext,above= of frame2]  {\scriptsize \vphantom{p}Depth map\vphantom{p}};
\node[Mytext,above= of frame3]  {\scriptsize \vphantom{p}PixDepth\vphantom{p}};

\node[Mytext,above= of frame14]  {\scriptsize \vphantom{p}Input Image\vphantom{p}};
\node[Mytext,above= of frame15]  {\scriptsize \vphantom{p}Depth map\vphantom{p}};

\node[Mytext,above= of frame16]  {\scriptsize \vphantom{p}PixPro\vphantom{p}};
\node[Mytext,above= of frame17]  {\scriptsize \vphantom{p}PixDepth\vphantom{p}};

\end{tikzpicture}
}
    \vspace{-1em}
    \caption{Comparison of the quality of the learned representations. The pixels in the similarity maps correspond to the reference feature vector, which is compared to all the rest of the vectors forming the features map. Note that these images are part of the validation set (unseen during training), and that the depth map is given for comparison purposes, it is not fed into the encoder.}
    \label{fig:repsim}
\end{figure*}
\paragraph{Implementation details}
\vspace{-0.2cm}
In the pre-training stage we used the SGD optimizer with a momentum of 0.9, a linear warm-up of 20 epochs and a cosine annealing scheduler for the rest of the training. We also fixed the maximum learning rate at $0.1$ and trained for a total 500 epochs and a batch size of 256 using 8 NVIDIA V100 GPUs for ImageNet and ScanNet, and 800 epochs and a batch size of 128 using 2 NVIDIA V100s for US BoreholeImage. 
In the fine-tuning stage we trained for 80 epochs with an early stopping using the Adam optimizer with a base learning rate of $0.001$ and we used a cosine annealing scheduler. We fixed the batch size at 32 for all the segmentation tasks.

\vspace{-0.45cm}
\paragraph{Representation Quality Evaluation}
To visually evaluate and compare the quality of the learned representations with PixDepth and PixPro, we propose the setup shown in Figure~\ref{fig:qlttveexp}. 
We take an input image and we resize it to a large size (\textit{e.g.} $896\times 896$) so that the output feature map $\mathcal{F}$ can be large enough (\textit{e.g.} $32\times 32$). Each vector of this map represents a patch in the original image (the size of this patch being approximately the size of the network's receptive field). We then select a feature vector $\mathcal{F}_{i,j}$ corresponding to a distinct area in the image (\textit{e.g.} the head of a the cat) and we measure the cosine similarity of this vector with all the other vectors in $\mathcal{F}$ and we build a similarity map. Some examples from ImageNet are shown in Figure~\ref{fig:repsim} and in the Supplementary Material.
For PixDepth, we can see that the features belonging to the same object are more spatially coherent and more contrasted with the background. This indicates that feature vectors that constitute the objects are similar to each other and different from those representing the background. This experiment provides a qualitative view of the effect of using the depth prior in our method.

\begin{figure}
    \centering
    \includegraphics[width=\columnwidth]{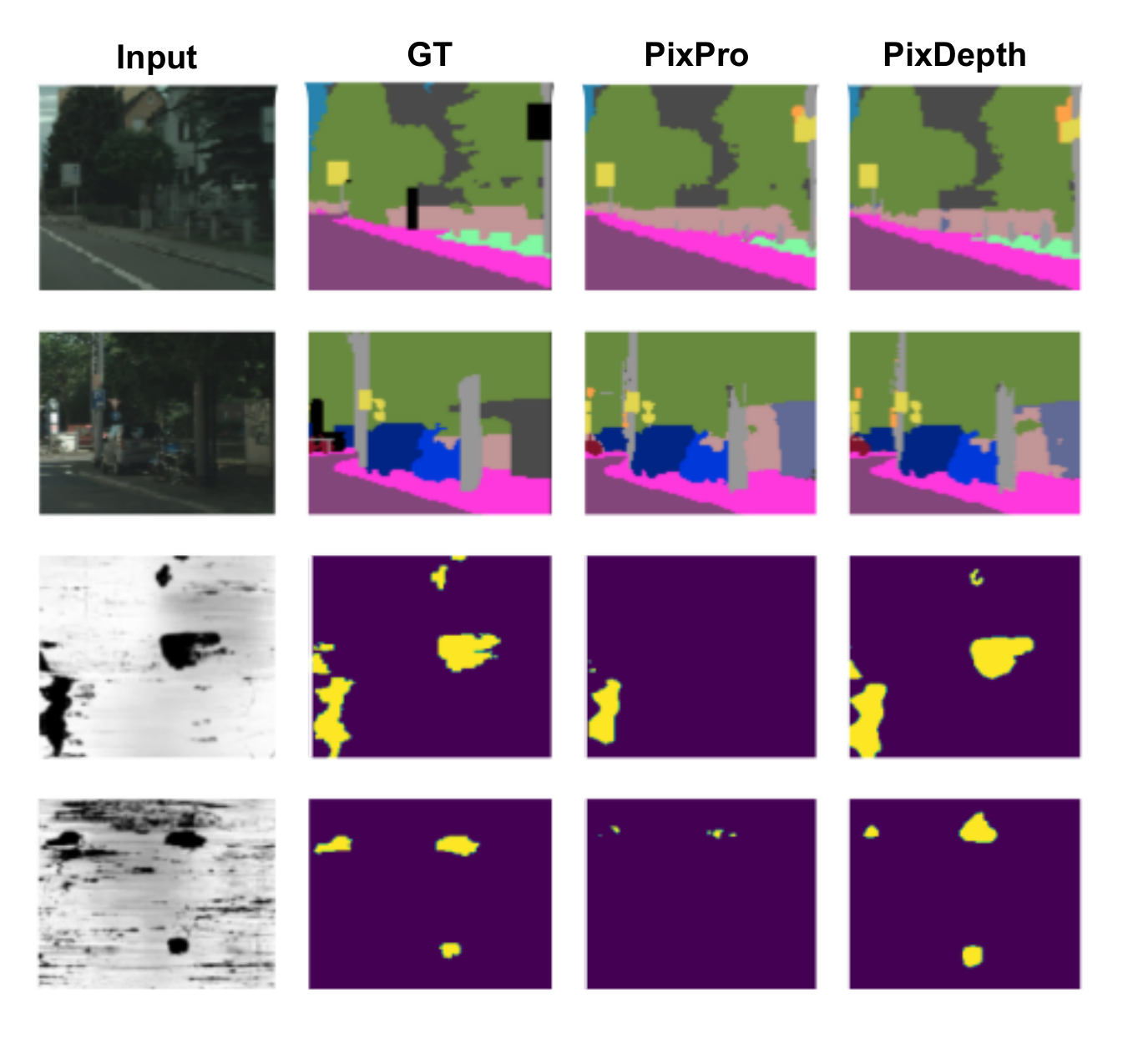}
    \vspace{-2.5em}
    \caption{Some qualitative examples for segmentation results on CityScapes (first and second rows) and Breakout Segmentation (third and fourth rows).}
    \label{fig:seg_examples}
\vspace{-0.40cm}
\end{figure}
\subsection{Results and Discussion}
Table~\ref{tab:results1} shows the results obtained on the breakout segmentation task. We observe that using the transit time with PixDepth improves the results over PixPro by 1.9\% and by 4.9\% over the supervised baseline. The transit time information helped the network to distinguish between real breakouts and imperfections in the images leading to better segmentation scores. 
Table~\ref{tab:results2} shows the results of Semantic Segmentation on CityScapes. Again, our method exhibits an improvement by 1.1\% over PixPro and more than 5\% over the supervised baseline. We also observe that the results with the change of the pre-training dataset and the exogenous data it comes with (a fraction of the MIDAS training set with ground truth disparity/depth maps) remained better than PixPro by 0.5\%. 
Table~\ref{tab:results3} compares our PixDepth pre-training with PixPro and Pri3D on ScanNet. We observe an improvement of mean IoU by 3.4\% over PixPro and 0.6\% over Pri3D. Apart from confirming the improvement over PixPro on another task, these results show that our method, which uses less trainable parameters and a simpler procedure, achieves comparable results to Pri3D. 
We show also in Figure \ref{fig:seg_examples} some of the examples where we can qualitatively see that our method performs better than PixPro. 
This supports our claim that the prior depth information used for pairs selection boosts the quality of learned representations. Additional experiments with larger models (ResNet-50) were not possible in our study, but can be considered for future work.

\subsection{Ablation Study}
To demonstrate the impact of each part of the PixDepth, we conducted an ablation study on the multi-thresholding for $\mathcal{T}_i$ and $\mathcal{T'}_i$ on the BoreholeImage dataset. Results are shown in Table~\ref{tab:ablation}. This suggests that multi-thresholding does improve slightly the mIoU on the breakout segmentation task. But most importantly it gives a score closer to the ones obtained with the best parameters when set to a single threshold. However, using more than 3 thresholds gives worse results. This is likely due to the fact that using too many thresholds results in the division of the feature map into smaller fragments. These smaller features become less expressive and thus perform worse at the contrastive tasks.
\begin{table}[]
\centering
\begin{tabular}{ccc}
\hline
$\mathcal{T'}$ & $\mathcal{T}$ & mIoU \\ \hline
0.1 & 0.5 & 70.3 \\
0.3 &  & \textbf{74.0} \\
0.5 &  & 71.4 \\
0.7 &  & 69.8 \\ \hline
0.3 & 0.1 & 71.6 \\
 & 0.3 & 72.1 \\
 & 0.5 & \textbf{74.0} \\
 & 0.7 & 72.4 \\
 & \{0.3, 0.7\} & 73.7 \\
 & \{0.3, 0.5, 0.7\} & 73.9 \\ \hline
\{0.3, 0.7\} & 0.5 & 72.6 \\
\{0.3, 0.5, 0.7\} &  & \textbf{74.4} \\
\{0.1, 0.3, 0.5, 0.7\} &  & 73.7 \\ \hline
\{0.3, 0.5, 0.7 \}& \{0.3, 0.5, 0.7\} & \underline{74.2} \\ \hline
\end{tabular}
\caption{The effect of varying the thresholds $\mathcal{T}$ and $\mathcal{T'}$ on the segmentation performances on the breakout detection task. Multiple values indicate multi-thresholding.}
\label{tab:ablation}
\vspace{-0.4cm}
\end{table}

\section{Conclusion}
We have introduced PixDepth, an improvement to an existing CL based method for pixel-level representation learning. The main goal is to leverage prior depth information, which can be inferred using a pre-trained monocular depth estimation network or calculated/collected along with the images. We show that this idea along with the use of multiple thresholds (to take into account different scales in learned representations) results in better segmentation scores and improves the coherence of learned representations. We have supported our findings by multiple experiments in different datasets such as segmentation of indoor scenes on CityScapes and ScanNet and geological features identification on BoreholeImage dataset. Future work will focus on exploring the use PixDepth with more modern backbone architectures (based on transformers) and its impact on the pixel-level downstream tasks.

\ifwacvfinal
\vspace{-0.5cm}
\paragraph{Acknowledgments}
Contains information provided by the Oil and Gas Authority and/or other third parties.
We acknowledge support from ANRT CIFRE Ph.D. scholarship n$^\circ$2020/0153 of the MESRI. This work was performed using HPC resources 
from GENCI–IDRIS (grant 2022-AD011011801R2) and from the “Mésocentre” computing center of CentraleSupélec and ENS 
Paris-Saclay supported by CNRS and Région Île-de-France (http://mesocentre.centralesupelec.fr/).
\fi

{\small
\bibliographystyle{ieee_fullname}
\bibliography{egpaper_final}
}

\end{document}


\maketitle
\section*{More results for feature quality comparison}

\begin{figure*}
    \centering
    \resizebox{\linewidth}{!}{
    \begin{tikzpicture}[
 image/.style = {text width=0.1\textwidth,  inner sep=0pt, outer sep=0pt},
node distance = 1mm and 1mm
                        ] 
\node [image] (frame1)
    {\includegraphics[width=\linewidth]{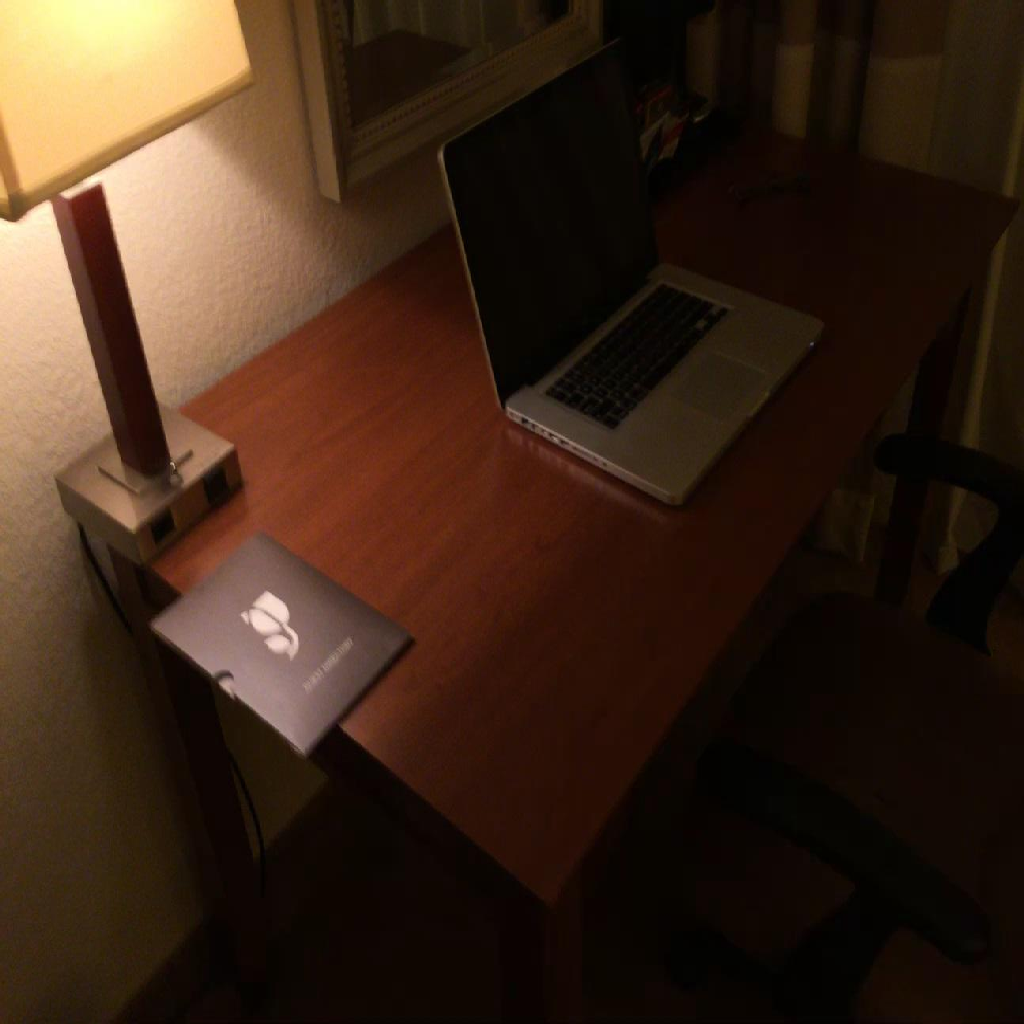}};
    
\node [image,right=of frame1] (frame2) 
    {\includegraphics[width=\linewidth,height=\linewidth]{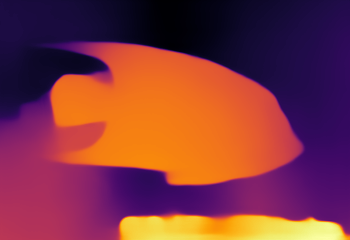}};
\node [image,right=of frame2] (frame222) 
    {\includegraphics[width=\linewidth]{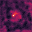}};
\node [image,right=of frame222] (frame3) 
    {\includegraphics[width=\linewidth]{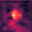}};
\node[image,below=of frame1] (frame4)
    {\includegraphics[width=\linewidth]{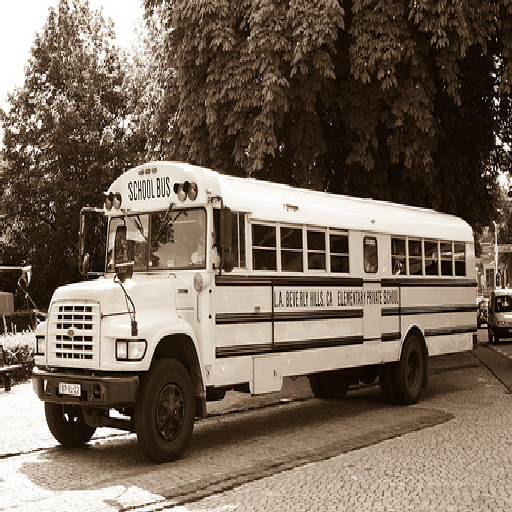}};
\node[image,right=of frame4] (frame44)
{\includegraphics[width=\linewidth,height=\linewidth]{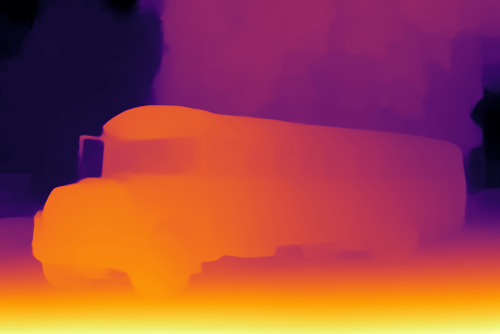}};
\node[image,right=of frame44] (frame5)
    {\includegraphics[width=\linewidth,height=\linewidth]{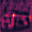}};
\node[image,right=of frame5] (frame6)
    {\includegraphics[width=\linewidth,height=\linewidth]{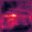}};
\node[image,below=of frame4] (frame7)
    {\includegraphics[width=\linewidth,height=\linewidth]{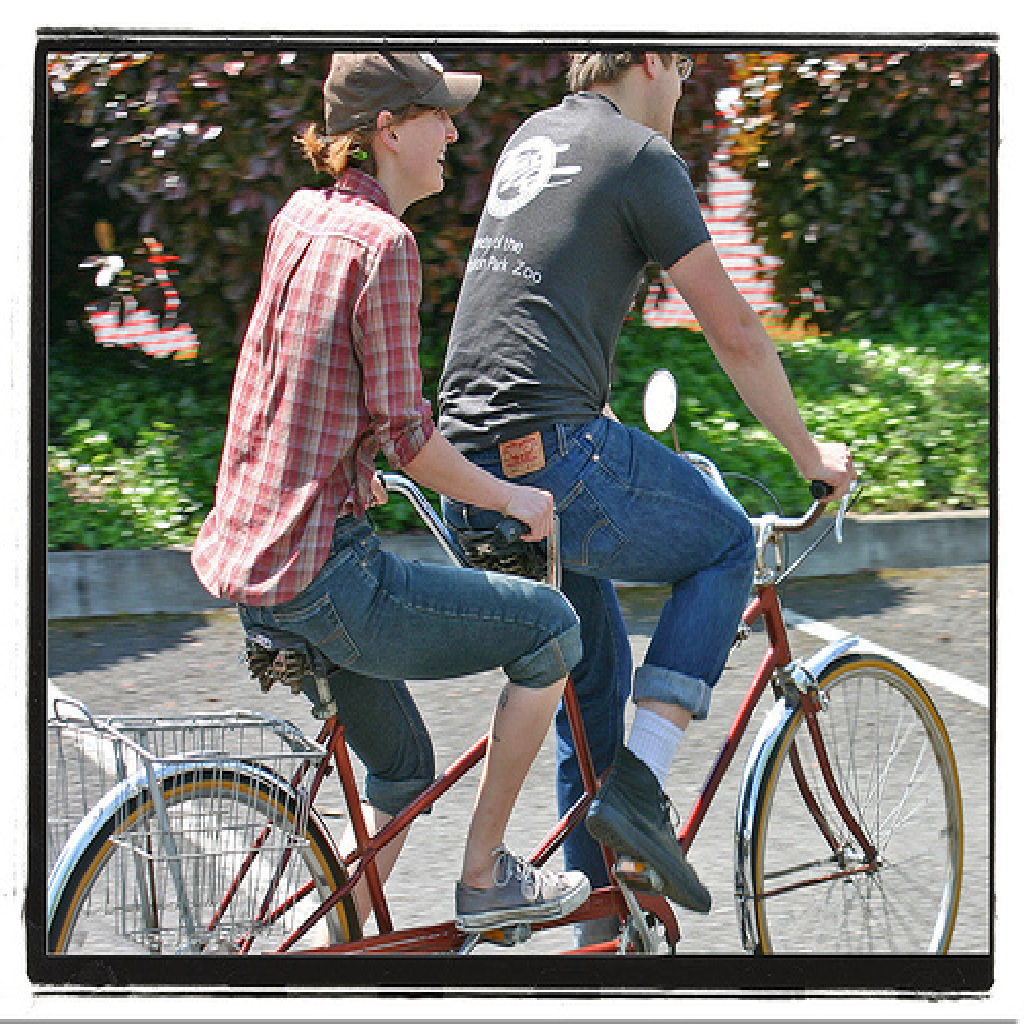}};
\node[image,right=of frame7] (frame77)
    {\includegraphics[width=\linewidth, height=\linewidth]{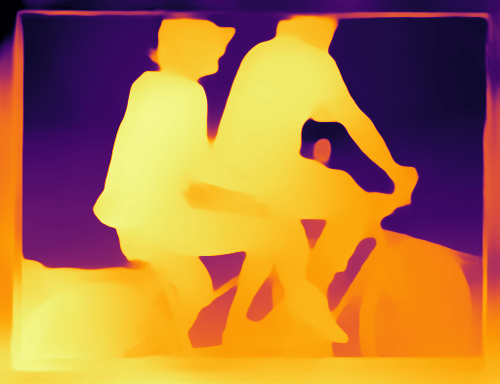}};
\node[image,right=of frame77] (frame8)
    {\includegraphics[width=\linewidth,height=\linewidth]{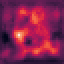}};
\node[image,right=of frame8] (frame9)
    {\includegraphics[width=\linewidth,height=\linewidth]{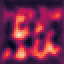}};

\node[image,right=0.2cm of frame3] (frame14)
    {\includegraphics[width=\linewidth,height=\linewidth]{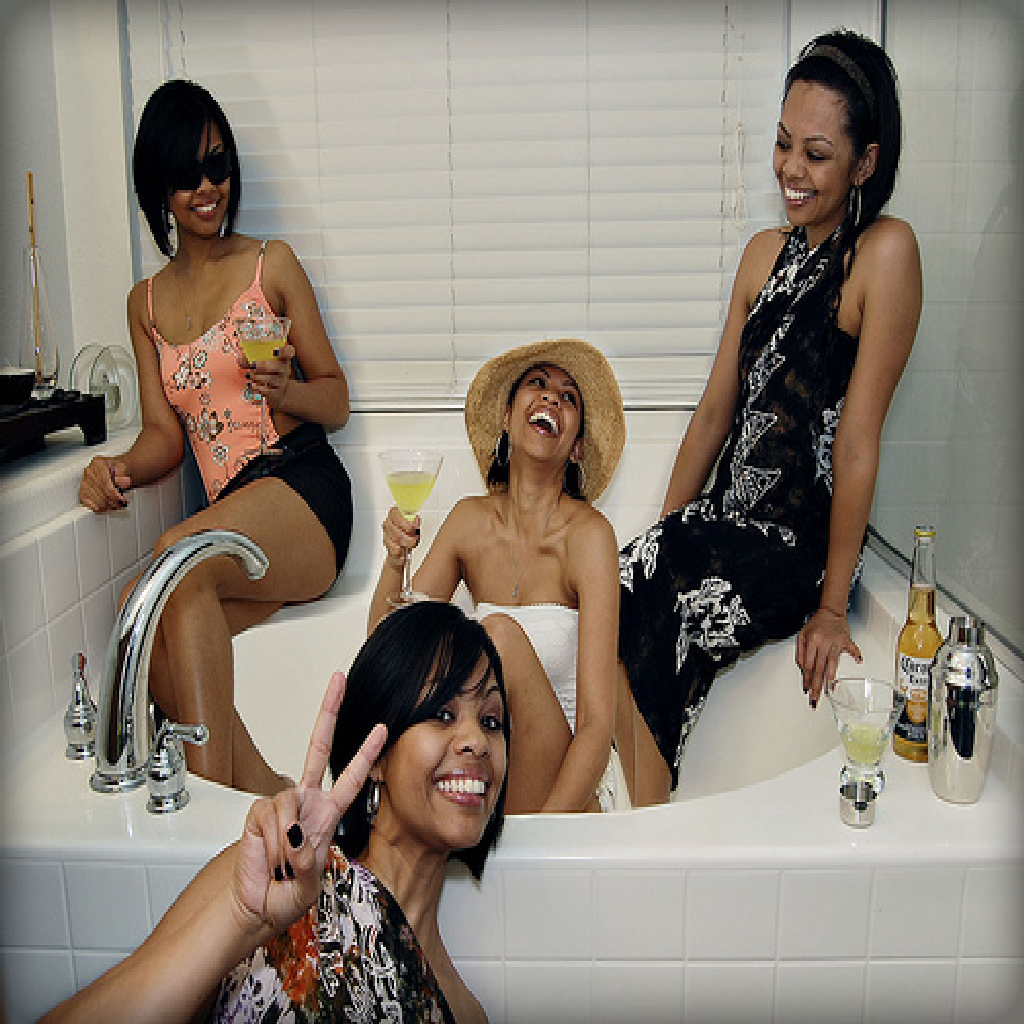}};
\node[image,right=of frame14] (frame15)
    {\includegraphics[width=\linewidth, height=\linewidth]{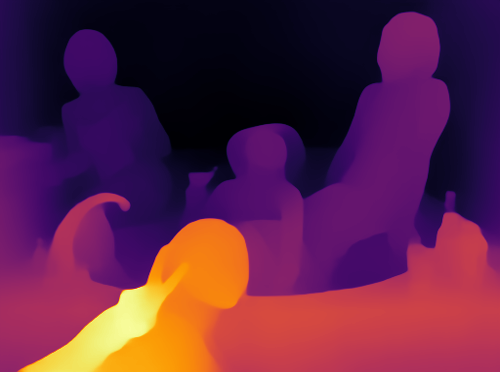}};
\node[image,right=of frame15] (frame16)
    {\includegraphics[width=\linewidth,height=\linewidth]{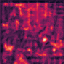}};
\node[image,right=of frame16] (frame17)
    {\includegraphics[width=\linewidth,height=\linewidth]{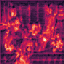}};

\node[image,below=of frame14] (frame10)
    {\includegraphics[width=\linewidth,height=\linewidth]{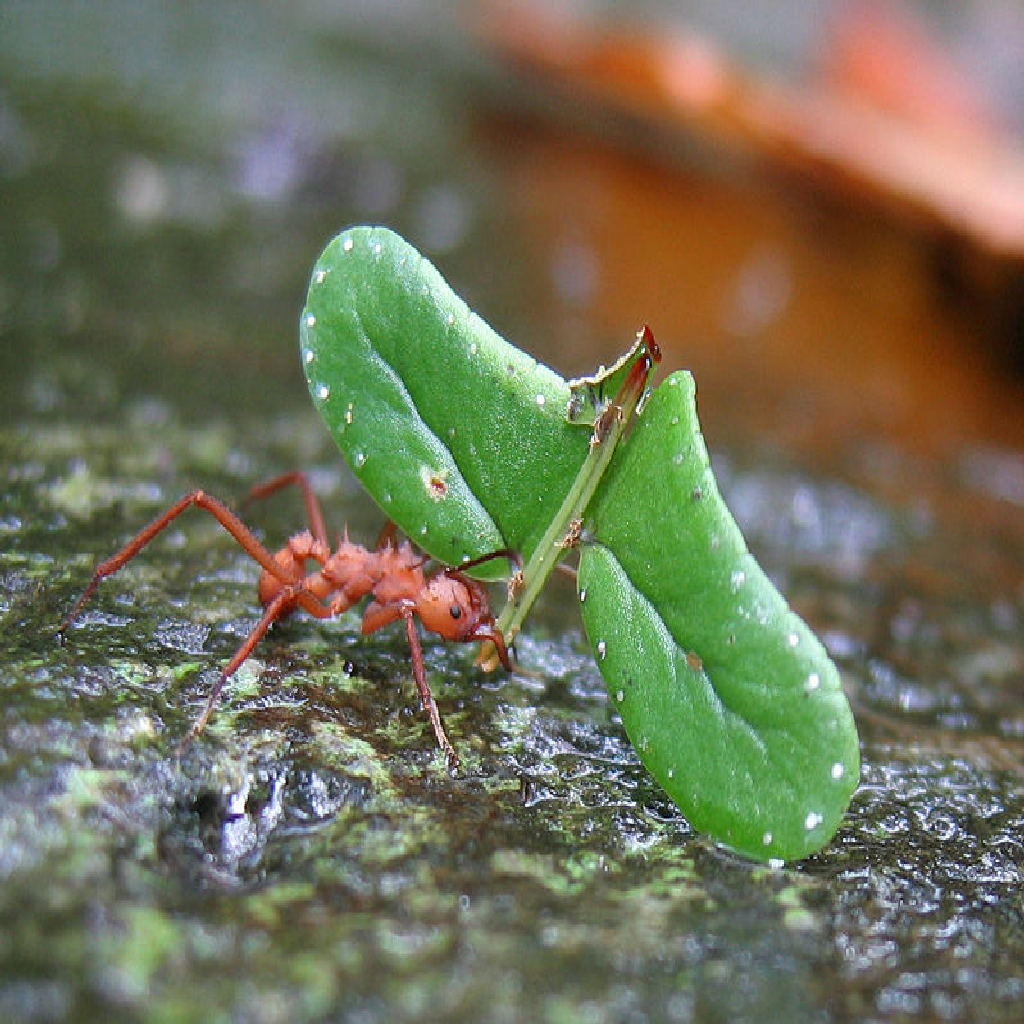}};
\node[image,right=of frame10] (frame11)
    {\includegraphics[width=\linewidth, height=\linewidth,height=\linewidth]{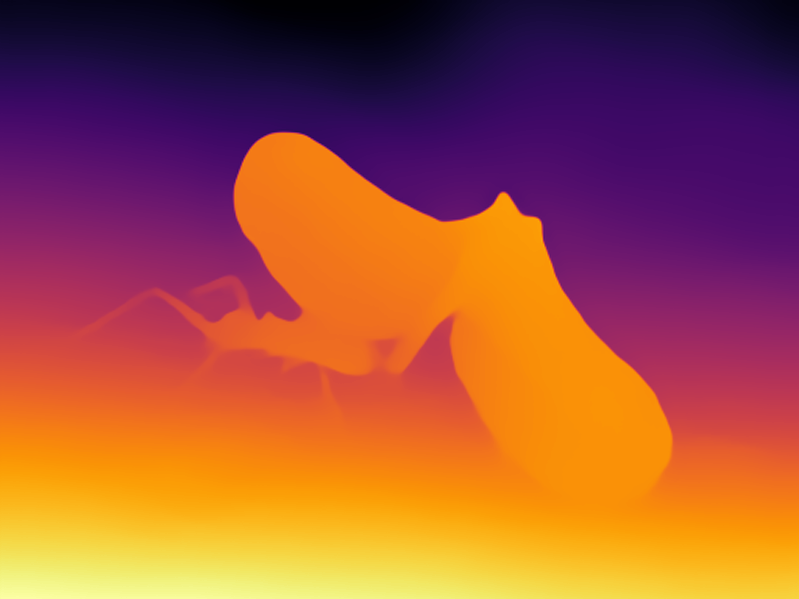}};
\node[image,right=of frame11] (frame12)
    {\includegraphics[width=\linewidth,height=\linewidth]{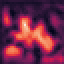}};
\node[image,right=of frame12] (frame13)
    {\includegraphics[width=\linewidth,height=\linewidth]{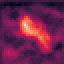}};

\node[image,below=of frame10] (frame18)
    {\includegraphics[width=\linewidth,height=\linewidth]{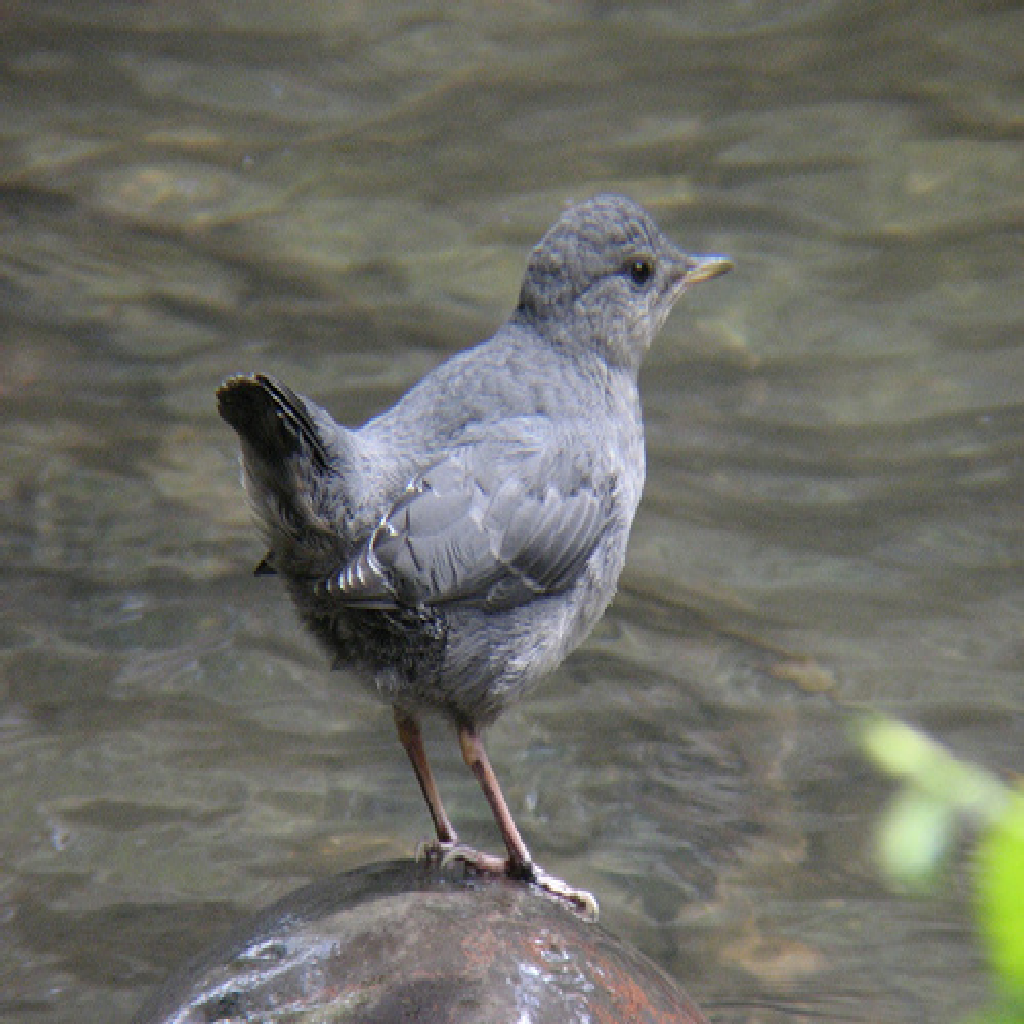}};
\node[image,right=of frame18] (frame19)
    {\includegraphics[width=\linewidth, height=\linewidth]{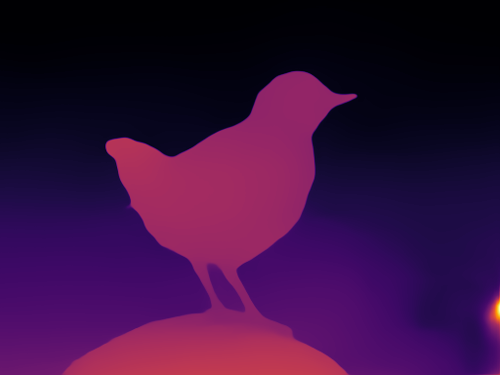}};
\node[image,right=of frame19] (frame20)
    {\includegraphics[width=\linewidth,height=\linewidth]{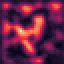}};
\node[image,right=of frame20] (frame21)
    {\includegraphics[width=\linewidth,height=\linewidth]{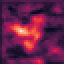}};

\node[image,below=of frame18] (frame50)
    {\includegraphics[width=\linewidth,height=\linewidth]{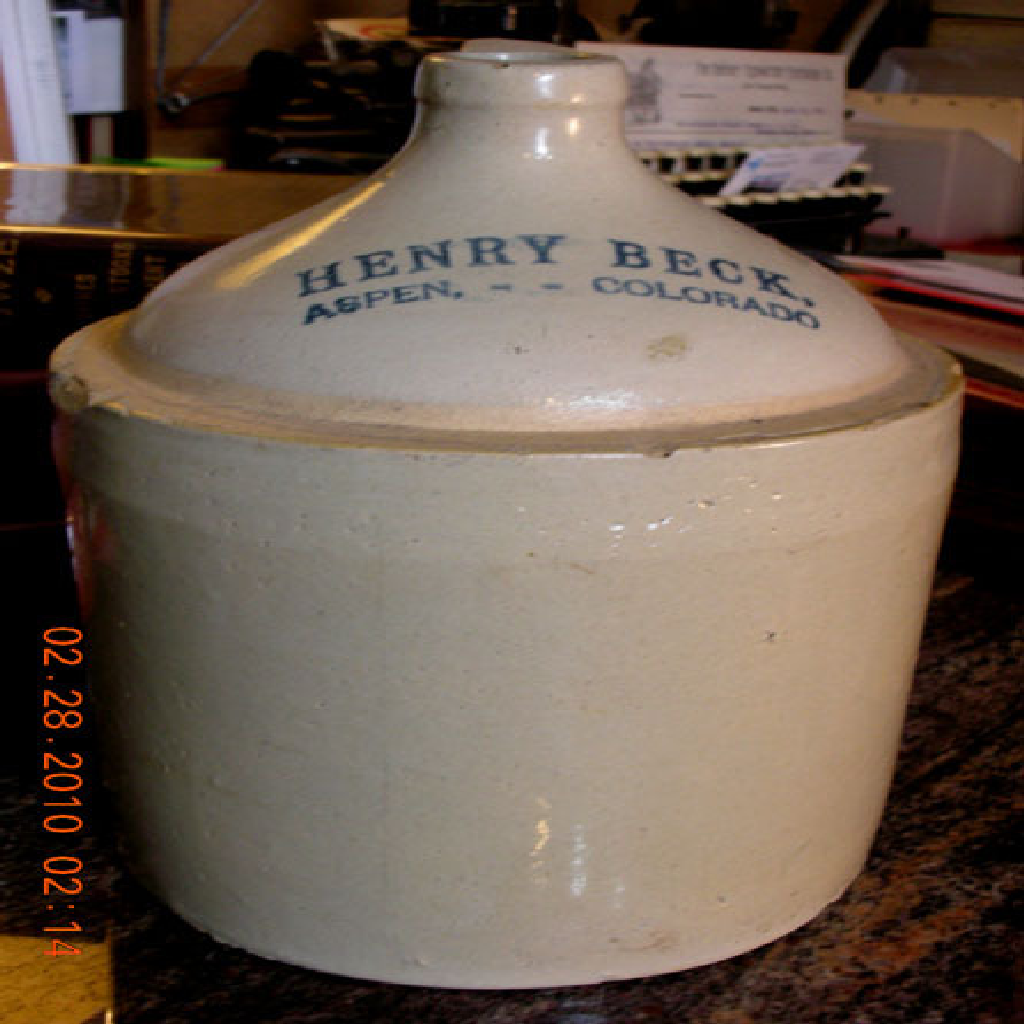}};
\node[image,right=of frame50] (frame51)
    {\includegraphics[width=\linewidth, height=\linewidth]{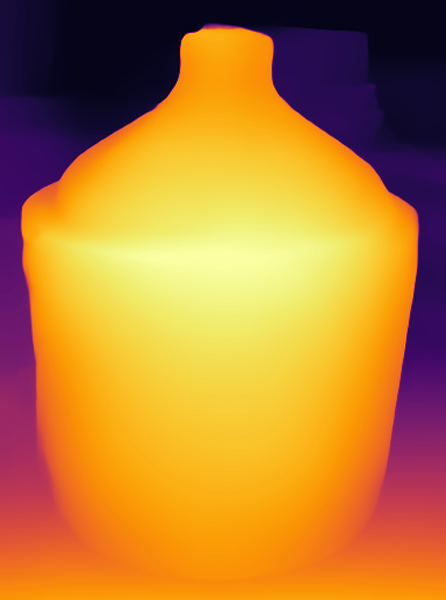}};
\node[image,right=of frame51] (frame52)
    {\includegraphics[width=\linewidth,height=\linewidth]{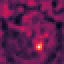}};
\node[image,right=of frame52] (frame53)
    {\includegraphics[width=\linewidth,height=\linewidth]{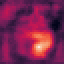}};

\node[image,below=of frame7] (frame54)
    {\includegraphics[width=\linewidth,height=\linewidth]{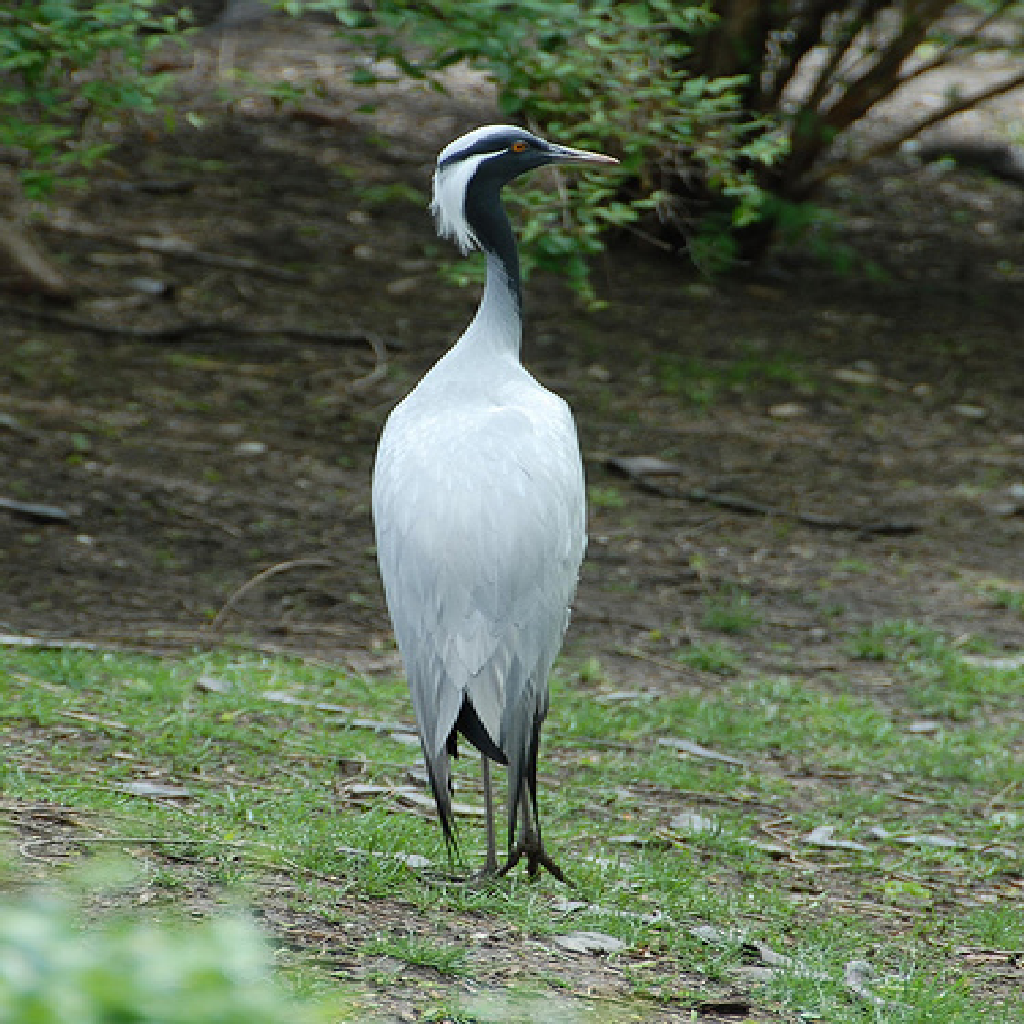}};
\node[image,right=of frame54] (frame55)
    {\includegraphics[width=\linewidth, height=\linewidth]{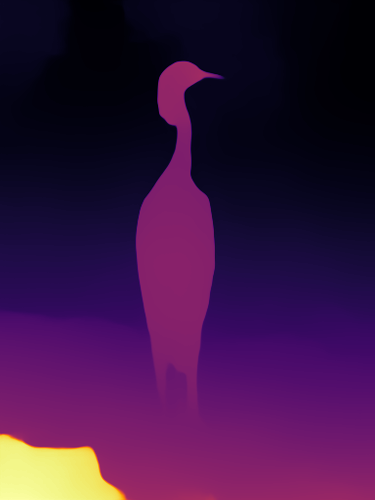}};
\node[image,right=of frame55] (frame56)
    {\includegraphics[width=\linewidth,height=\linewidth]{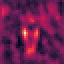}};
\node[image,right=of frame56] (frame57)
    {\includegraphics[width=\linewidth,height=\linewidth]{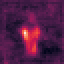}};

\node[text,above= of frame1]  {\scriptsize \vphantom{p}Input Image\vphantom{p}};
\node[text,above= of frame222]  {\scriptsize \vphantom{p}PixPro\vphantom{p}};

\node[text,above= of frame2]  {\scriptsize \vphantom{p}Depth map\vphantom{p}};
\node[text,above= of frame3]  {\scriptsize \vphantom{p}PixDepth\vphantom{p}};

\node[text,above= of frame14]  {\scriptsize \vphantom{p}Input Image\vphantom{p}};
\node[text,above= of frame15]  {\scriptsize \vphantom{p}Depth map\vphantom{p}};

\node[text,above= of frame16]  {\scriptsize \vphantom{p}PixPro\vphantom{p}};
\node[text,above= of frame17]  {\scriptsize \vphantom{p}PixDepth\vphantom{p}};

\end{tikzpicture}
}
    \vspace{-1em}
    \caption{
    More comparison of the quality of the learned representations on ImageNet images. The bright pixels in the similarity maps correspond to the reference feature vector, which is compared to all other vectors forming the features map. Note that these images are part of the validation set (unseen during training), and that the depth map is given for comparison purposes, it is not fed into the encoder.
    }
    \label{fig:results1}
\end{figure*}

\begin{figure*}
    \centering
    \resizebox{\linewidth}{!}{
    \begin{tikzpicture}[
 image/.style = {text width=0.1\textwidth,  inner sep=0pt, outer sep=0pt},
node distance = 1mm and 1mm
                        ] 
\node [image] (frame1)
    {\includegraphics[width=\linewidth]{figs_sup/21/input.png}};
    
\node [image,right=of frame1] (frame2) 
    {\includegraphics[width=\linewidth,height=\linewidth]{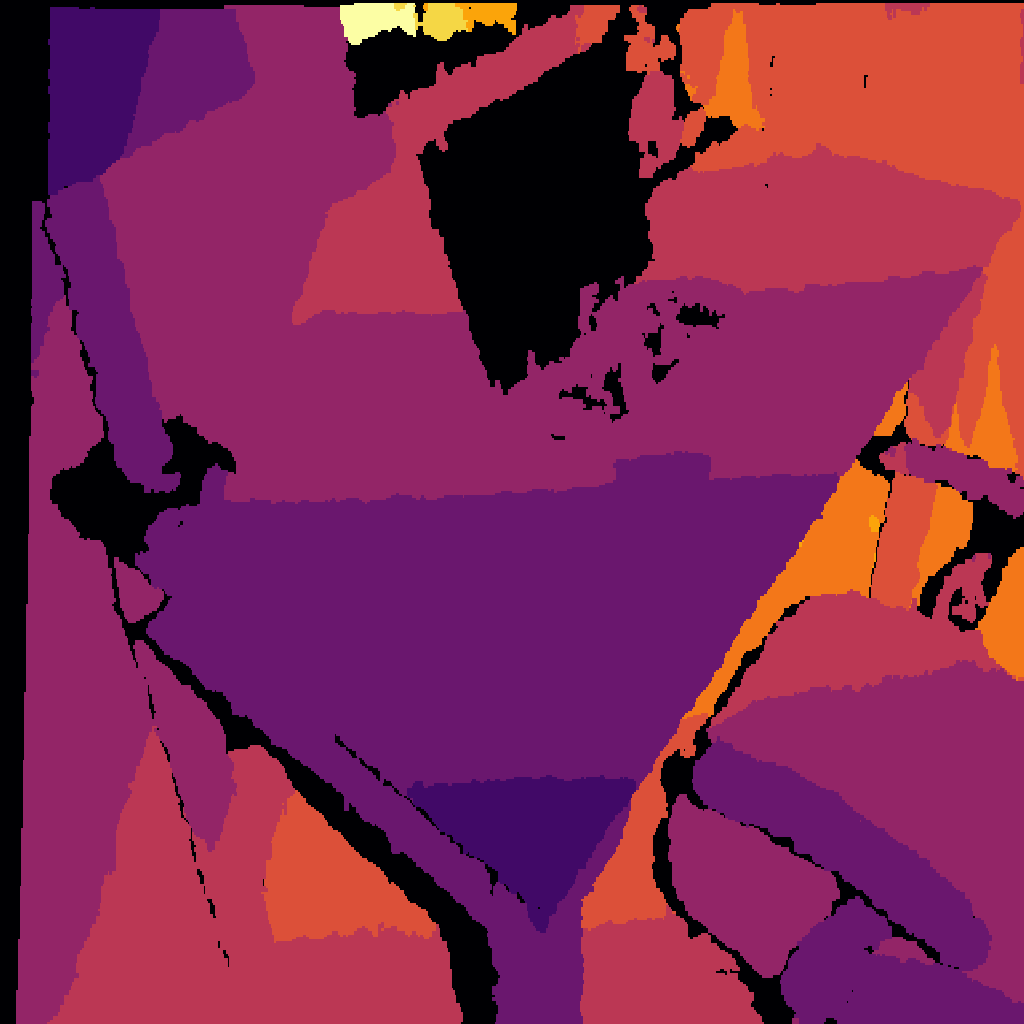}};
\node [image,right=of frame2] (frame222) 
    {\includegraphics[width=\linewidth]{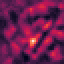}};
\node [image,right=of frame222] (frame3) 
    {\includegraphics[width=\linewidth]{figs_sup/11/pixelcl++.png}};
\node[image,below=of frame1] (frame4)
    {\includegraphics[width=\linewidth]{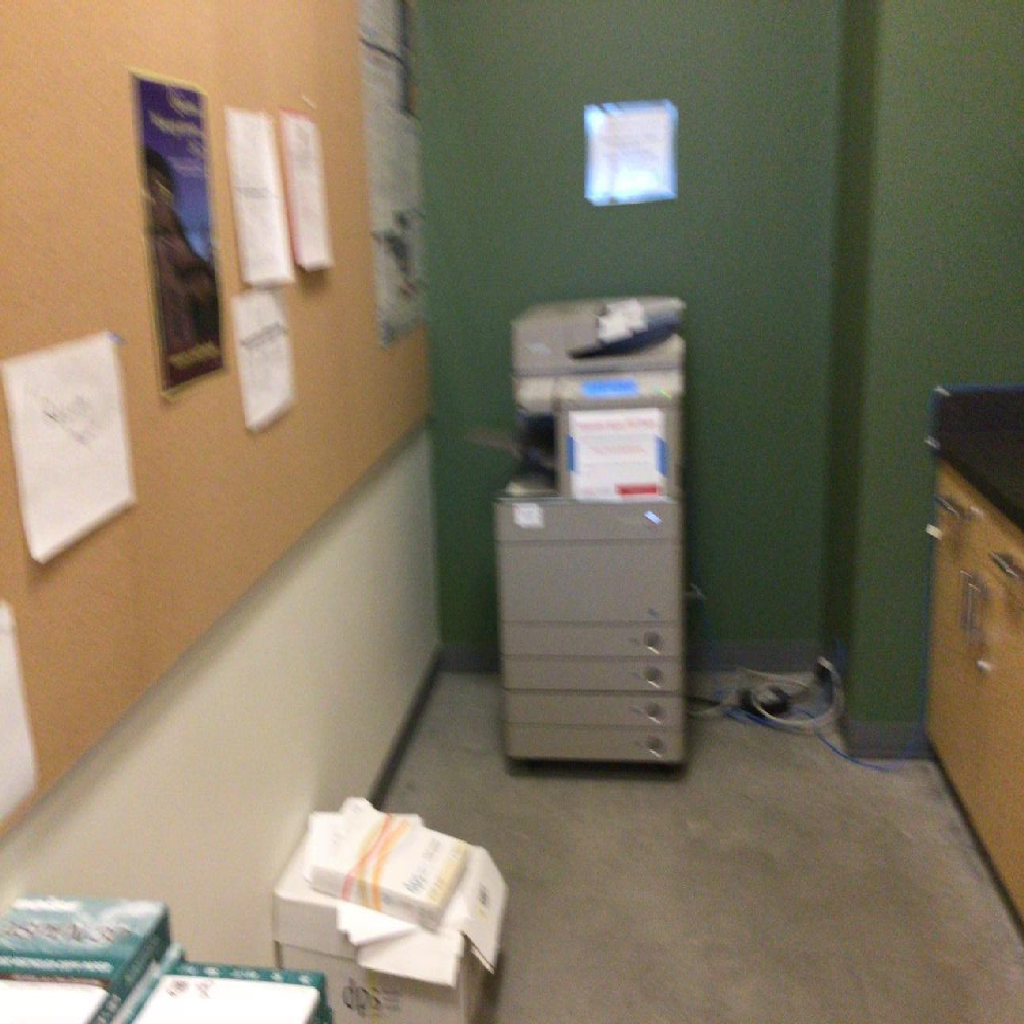}};
\node[image,right=of frame4] (frame44)
{\includegraphics[width=\linewidth,height=\linewidth]{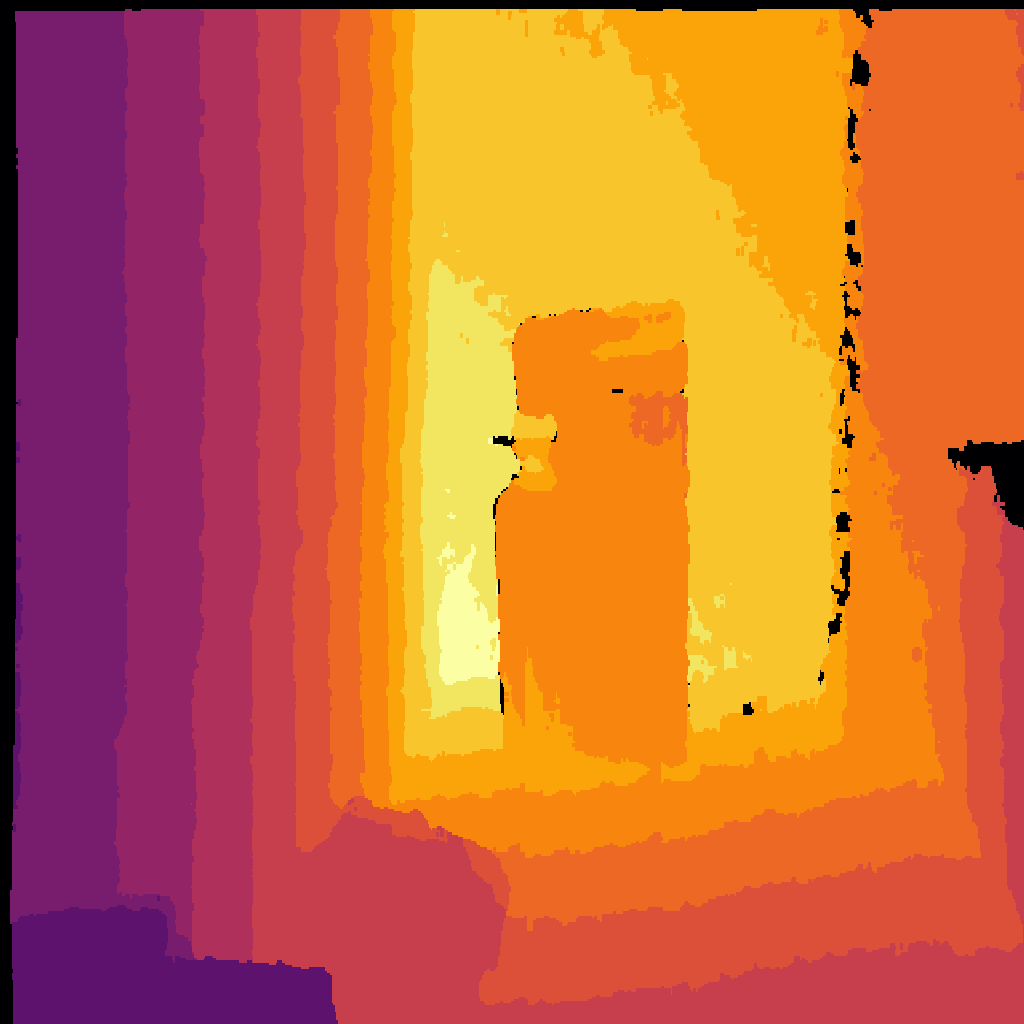}};
\node[image,right=of frame44] (frame5)
    {\includegraphics[width=\linewidth,height=\linewidth]{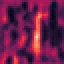}};
\node[image,right=of frame5] (frame6)
    {\includegraphics[width=\linewidth,height=\linewidth]{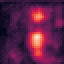}};
\node[image,below=of frame4] (frame7)
    {\includegraphics[width=\linewidth,height=\linewidth]{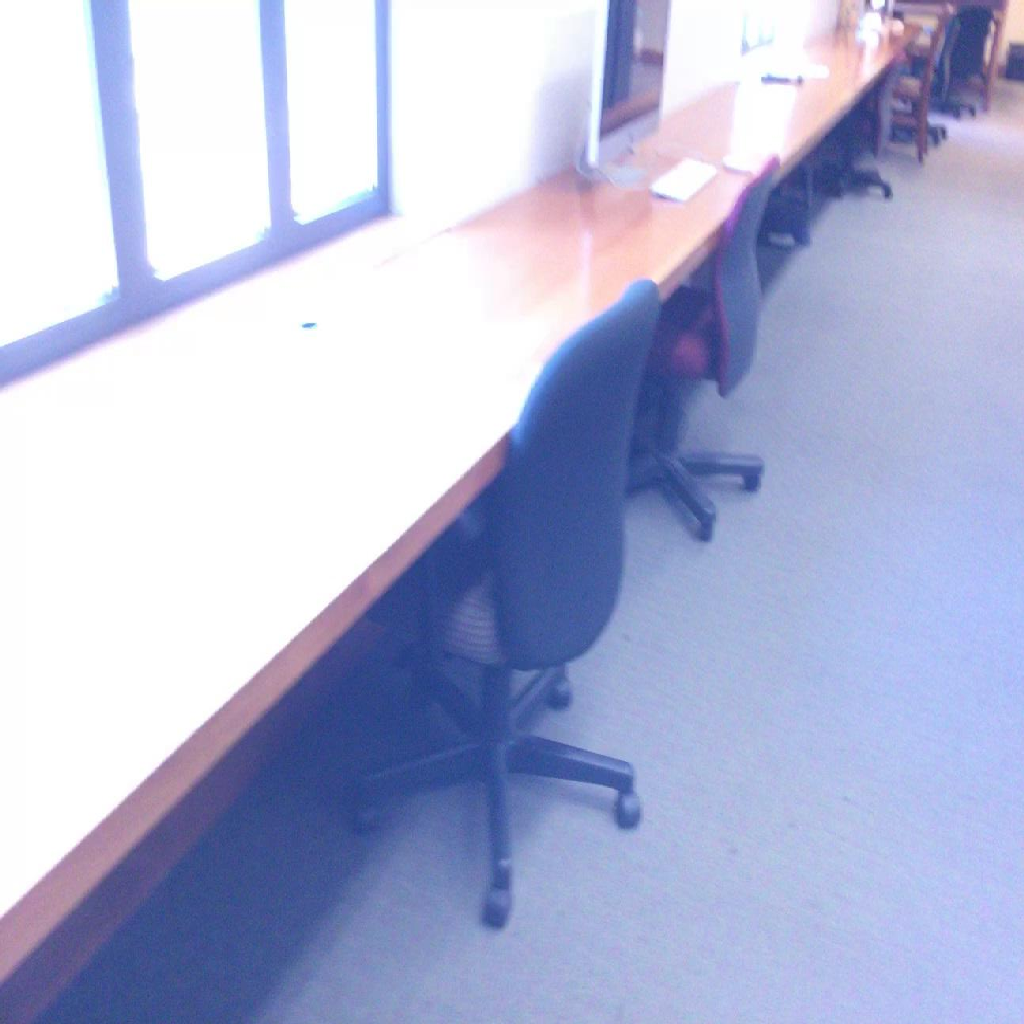}};
\node[image,right=of frame7] (frame777)
    {\includegraphics[width=\linewidth, height=\linewidth]{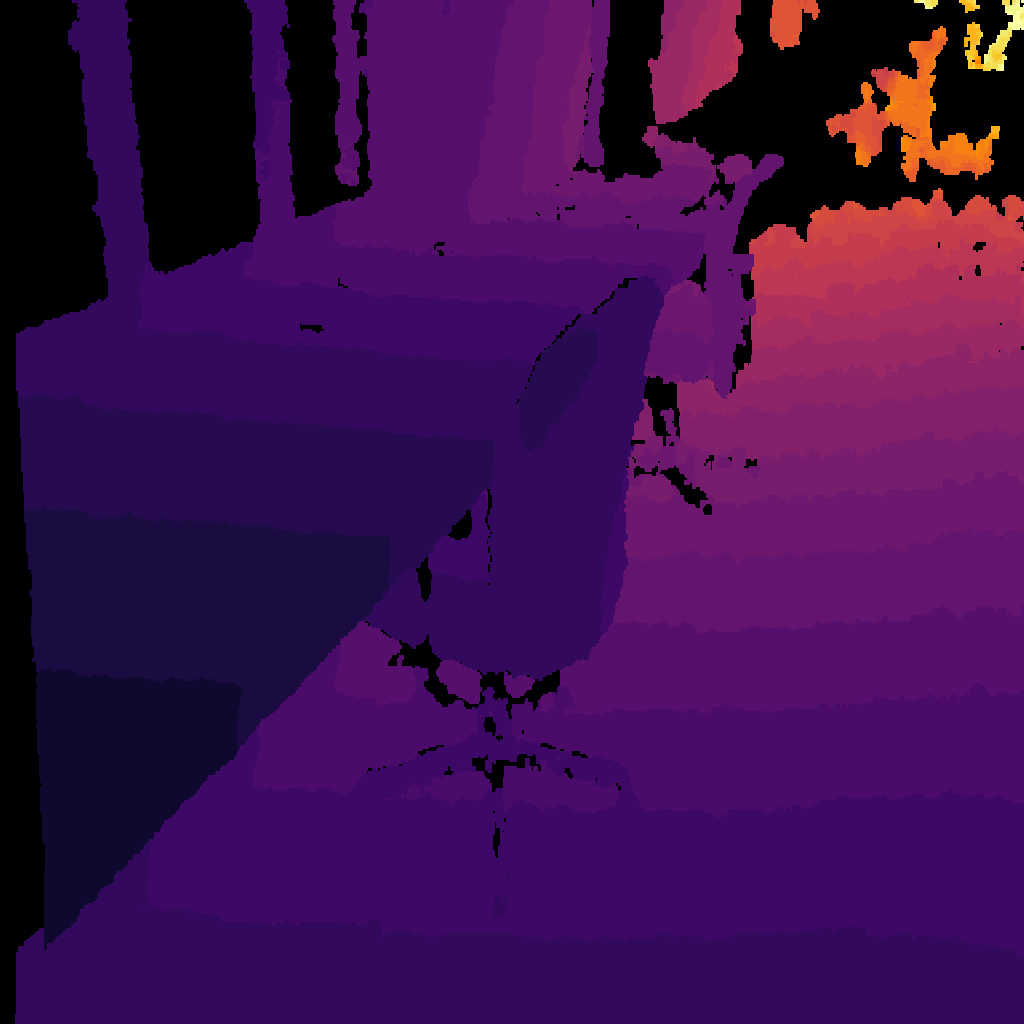}};
\node[image,right=of frame777] (frame8)
    {\includegraphics[width=\linewidth,height=\linewidth]{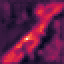}};
\node[image,right=of frame8] (frame9)
    {\includegraphics[width=\linewidth,height=\linewidth]{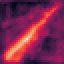}};

\node[image,right=0.2cm of frame3] (frame14)
    {\includegraphics[width=\linewidth,height=\linewidth]{figs_sup/23/input.png}};
\node[image,right=of frame14] (frame15)
    {\includegraphics[width=\linewidth, height=\linewidth]{figs_sup/23/depth.png}};
\node[image,right=of frame15] (frame16)
    {\includegraphics[width=\linewidth,height=\linewidth]{figs_sup/23/pixelcl.png}};
\node[image,right=of frame16] (frame17)
    {\includegraphics[width=\linewidth,height=\linewidth]{figs_sup/23/pixelcl++.png}};

\node[image,below=of frame14] (frame10)
    {\includegraphics[width=\linewidth,height=\linewidth]{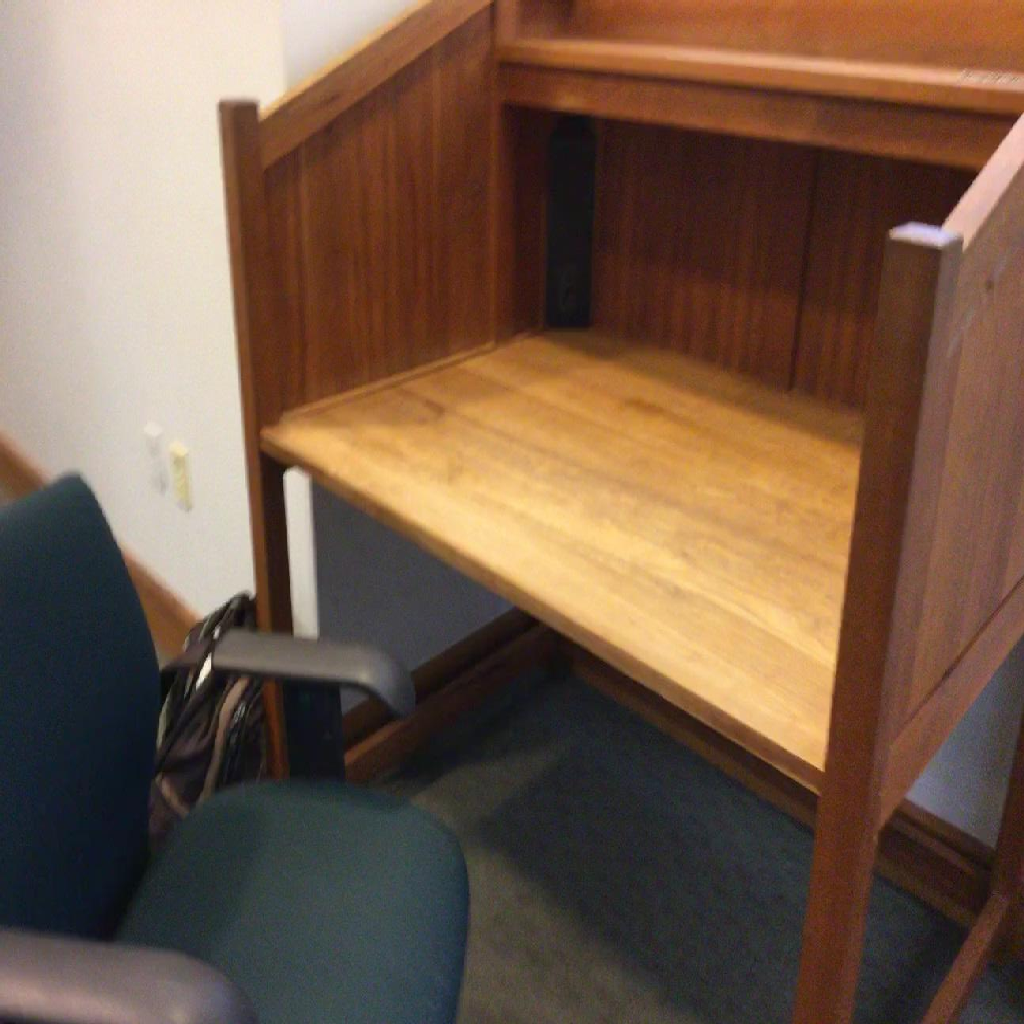}};
\node[image,right=of frame10] (frame11)
    {\includegraphics[width=\linewidth, height=\linewidth,height=\linewidth]{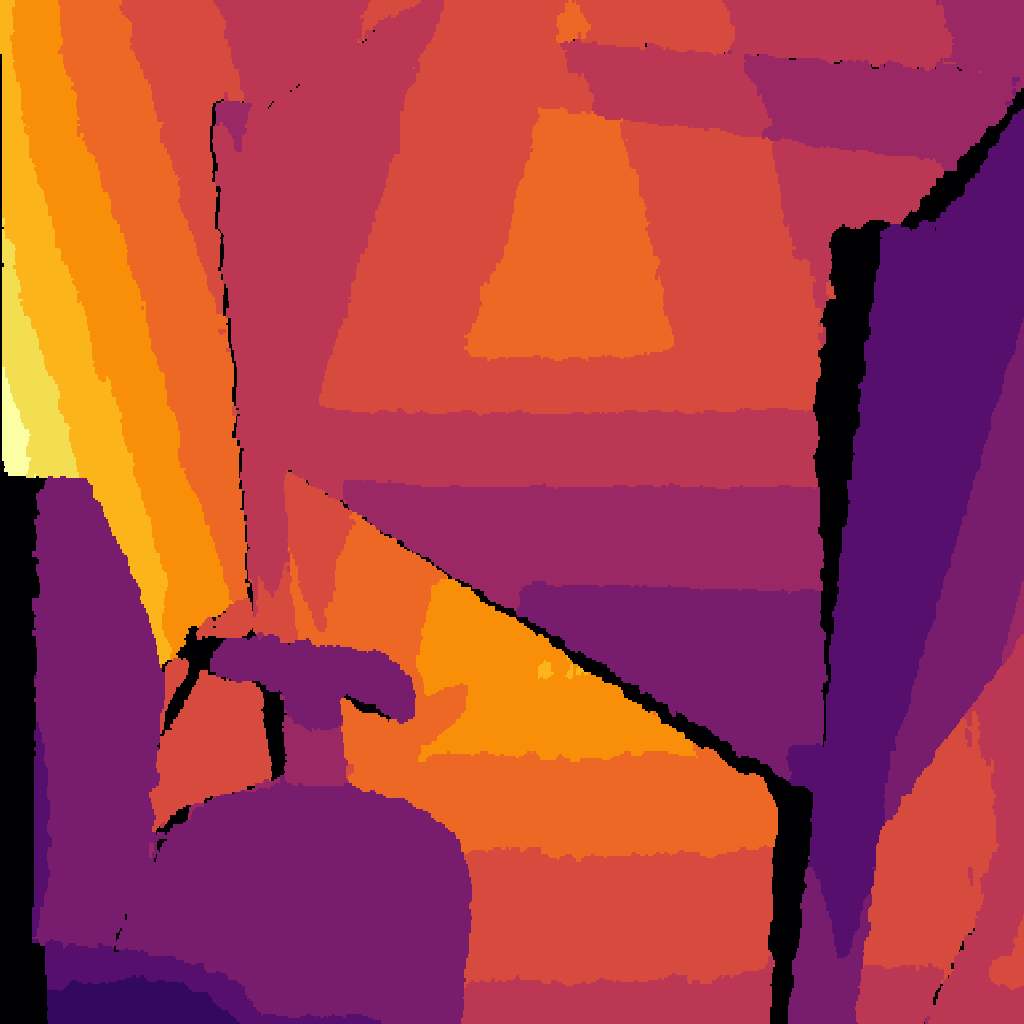}};
\node[image,right=of frame11] (frame12)
    {\includegraphics[width=\linewidth,height=\linewidth]{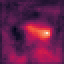}};
\node[image,right=of frame12] (frame13)
    {\includegraphics[width=\linewidth,height=\linewidth]{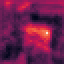}};

\node[image,below=of frame10] (frame18)
    {\includegraphics[width=\linewidth,height=\linewidth]{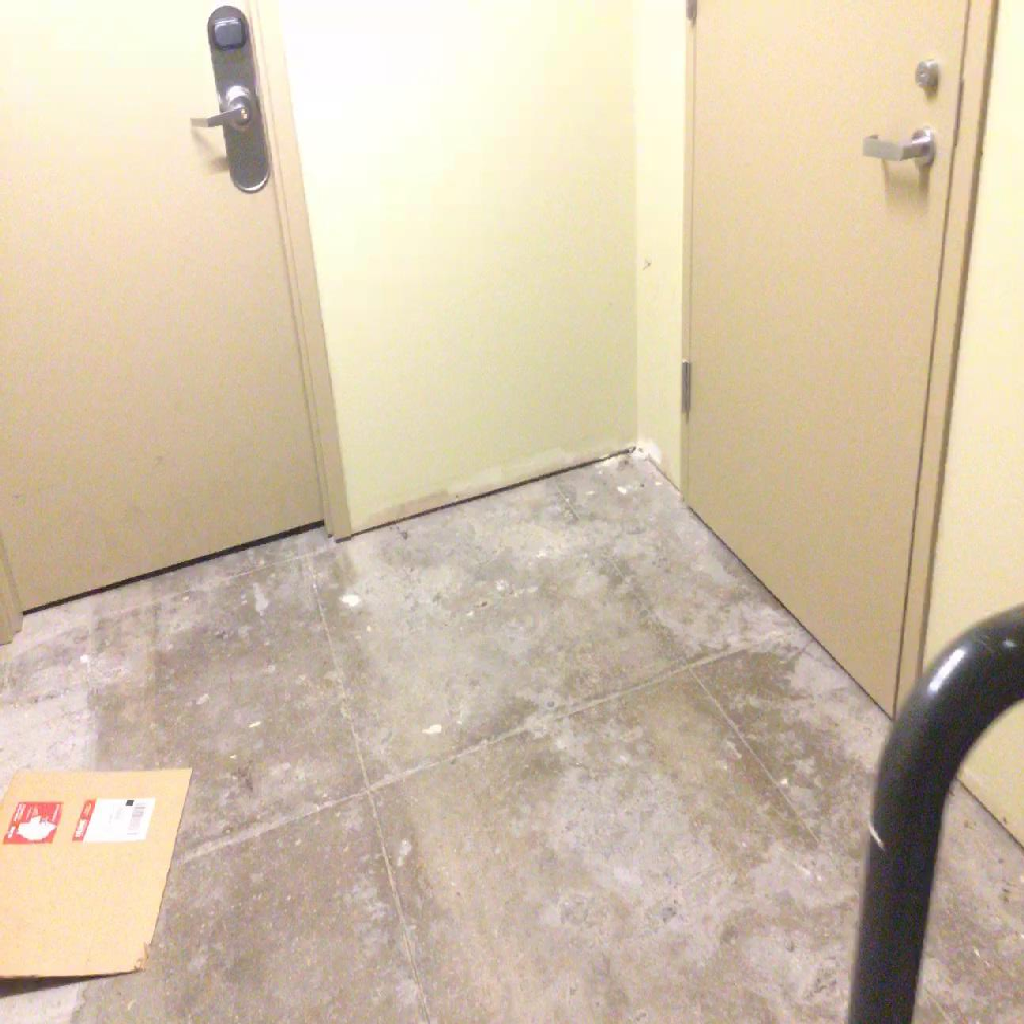}};
\node[image,right=of frame18] (frame19)
    {\includegraphics[width=\linewidth, height=\linewidth]{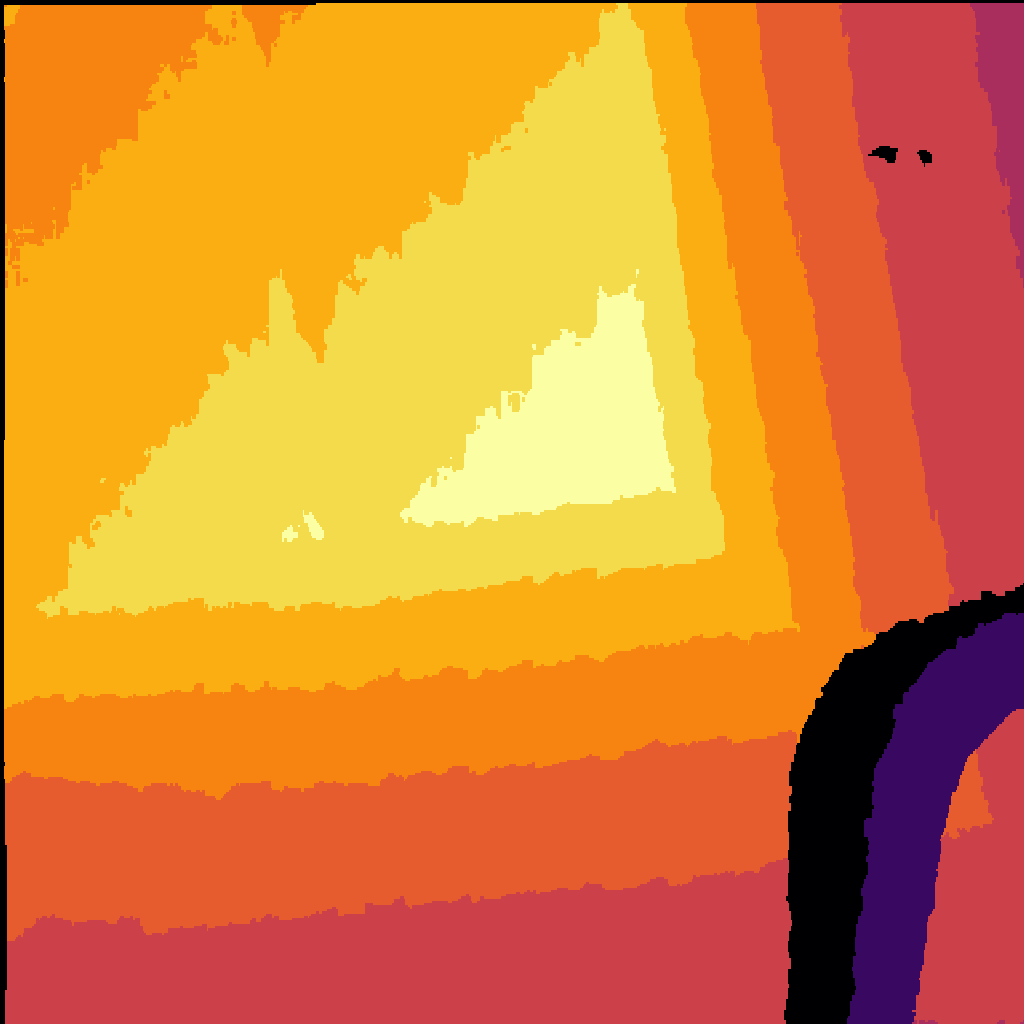}};
\node[image,right=of frame19] (frame20)
    {\includegraphics[width=\linewidth,height=\linewidth]{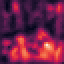}};
\node[image,right=of frame20] (frame21)
    {\includegraphics[width=\linewidth,height=\linewidth]{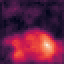}};

\node[image,below=of frame18] (frame50)
    {\includegraphics[width=\linewidth,height=\linewidth]{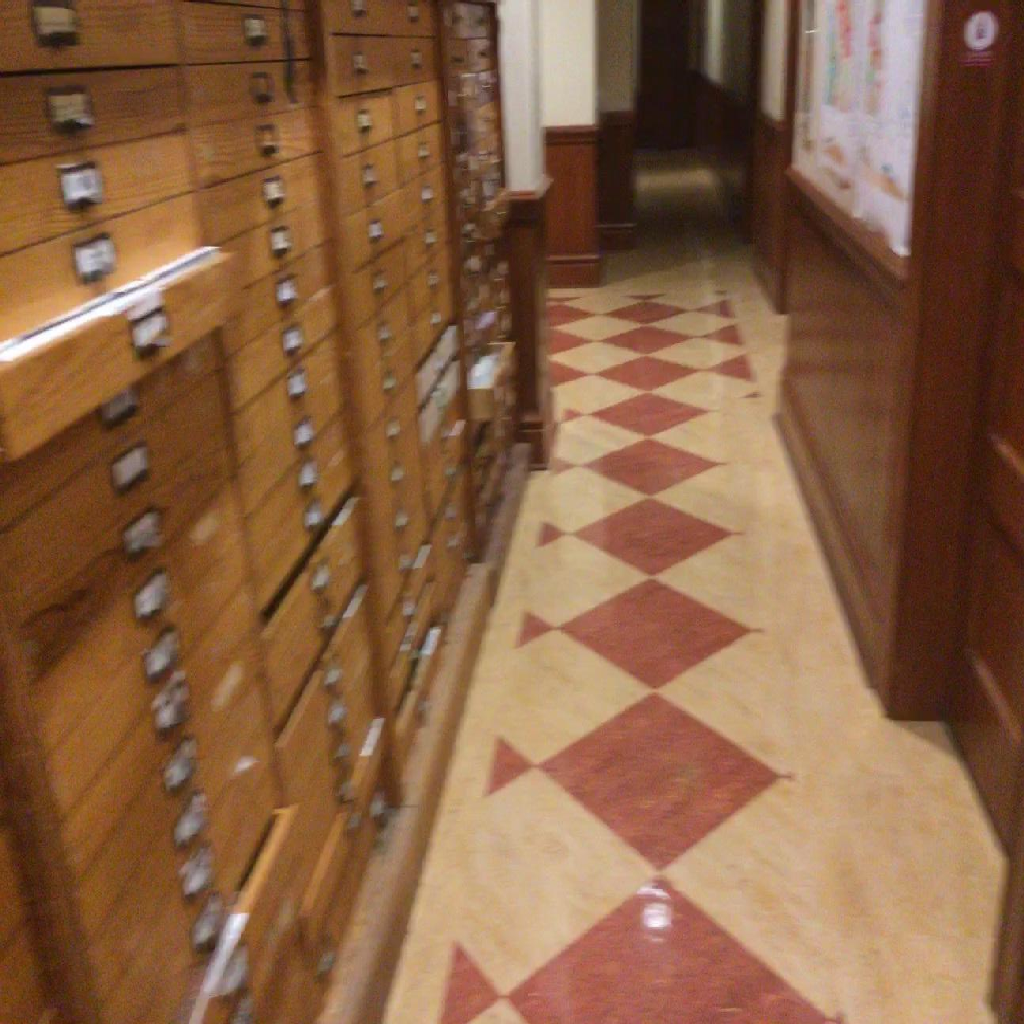}};
\node[image,right=of frame50] (frame51)
    {\includegraphics[width=\linewidth, height=\linewidth]{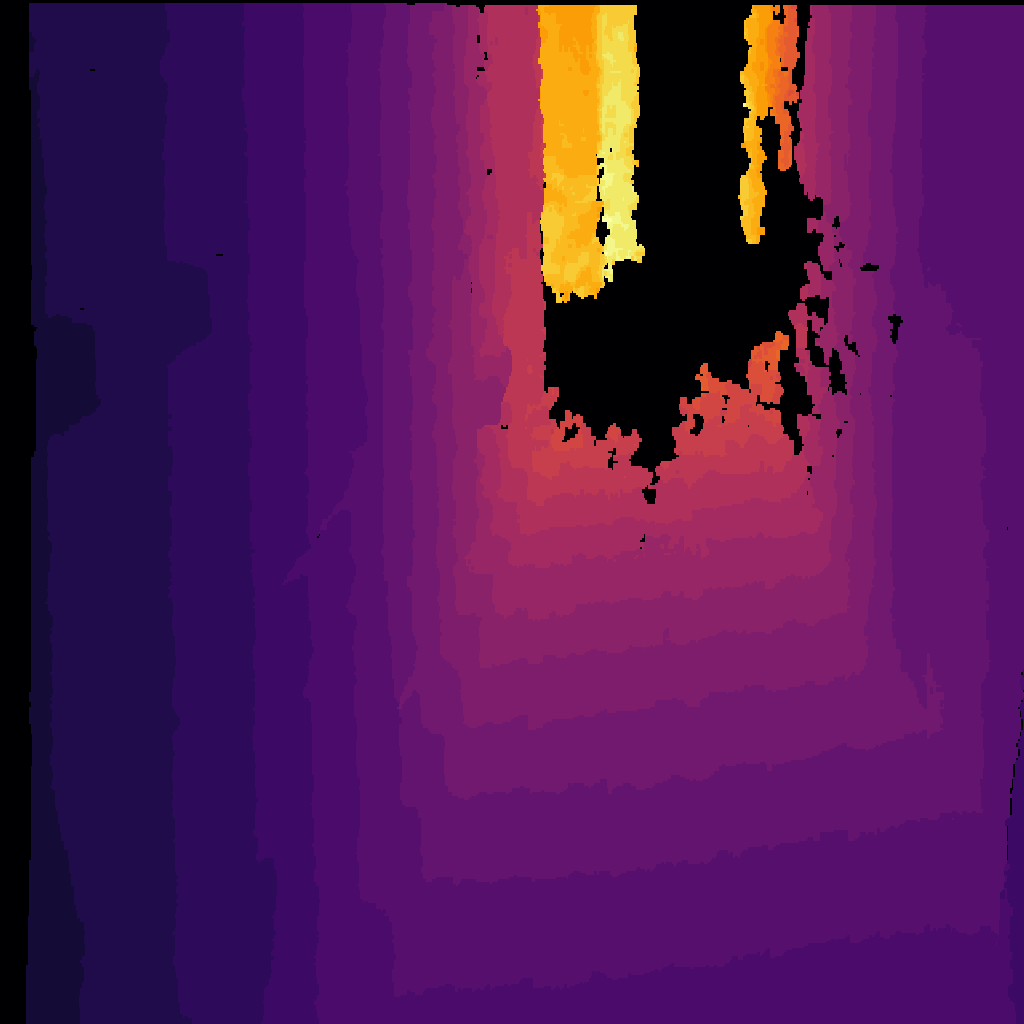}};
\node[image,right=of frame51] (frame52)
    {\includegraphics[width=\linewidth,height=\linewidth]{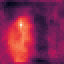}};
\node[image,right=of frame52] (frame53)
    {\includegraphics[width=\linewidth,height=\linewidth]{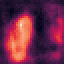}};

\node[image,below=of frame7] (frame54)
    {\includegraphics[width=\linewidth,height=\linewidth]{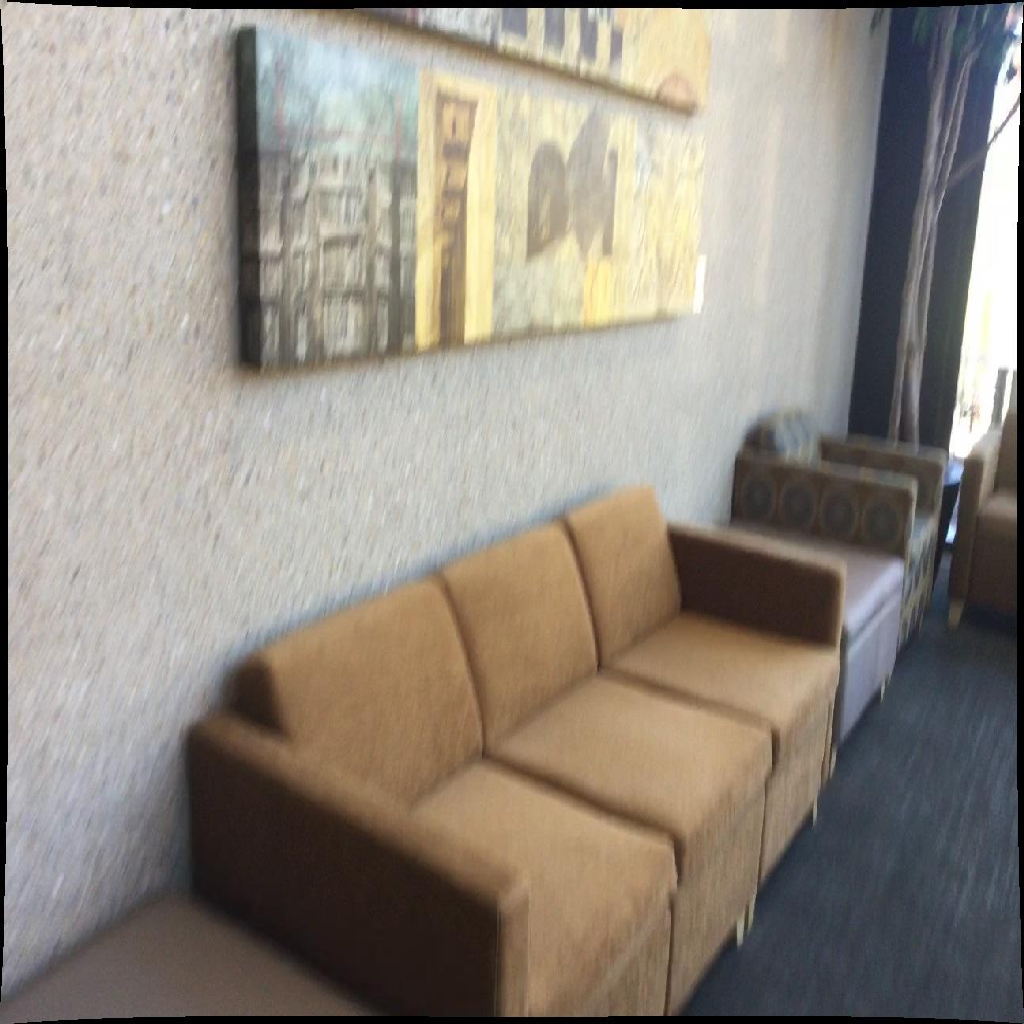}};
\node[image,right=of frame54] (frame55)
    {\includegraphics[width=\linewidth, height=\linewidth]{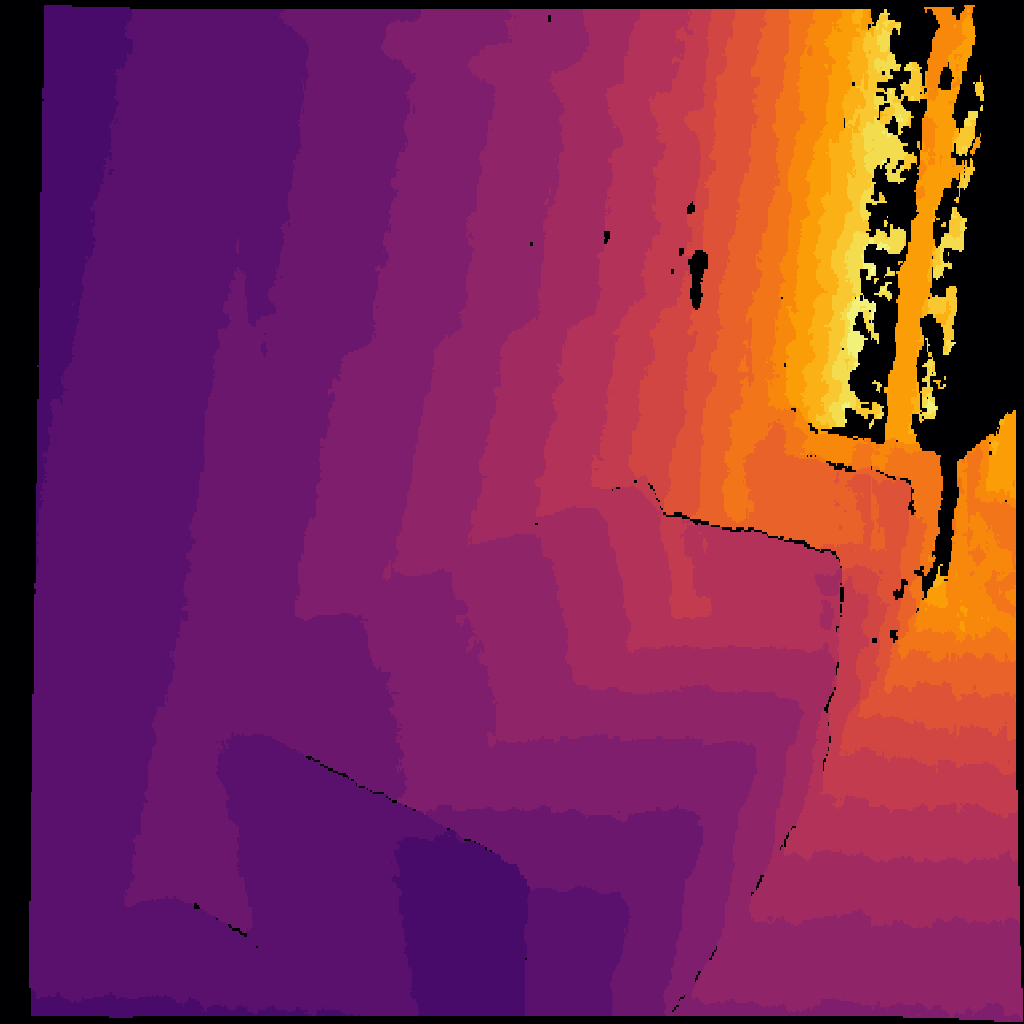}};
\node[image,right=of frame55] (frame56)
    {\includegraphics[width=\linewidth,height=\linewidth]{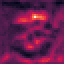}};
\node[image,right=of frame56] (frame57)
    {\includegraphics[width=\linewidth,height=\linewidth]{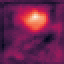}};

\node[image,below=of frame54] (frame58)
    {\includegraphics[width=\linewidth,height=\linewidth]{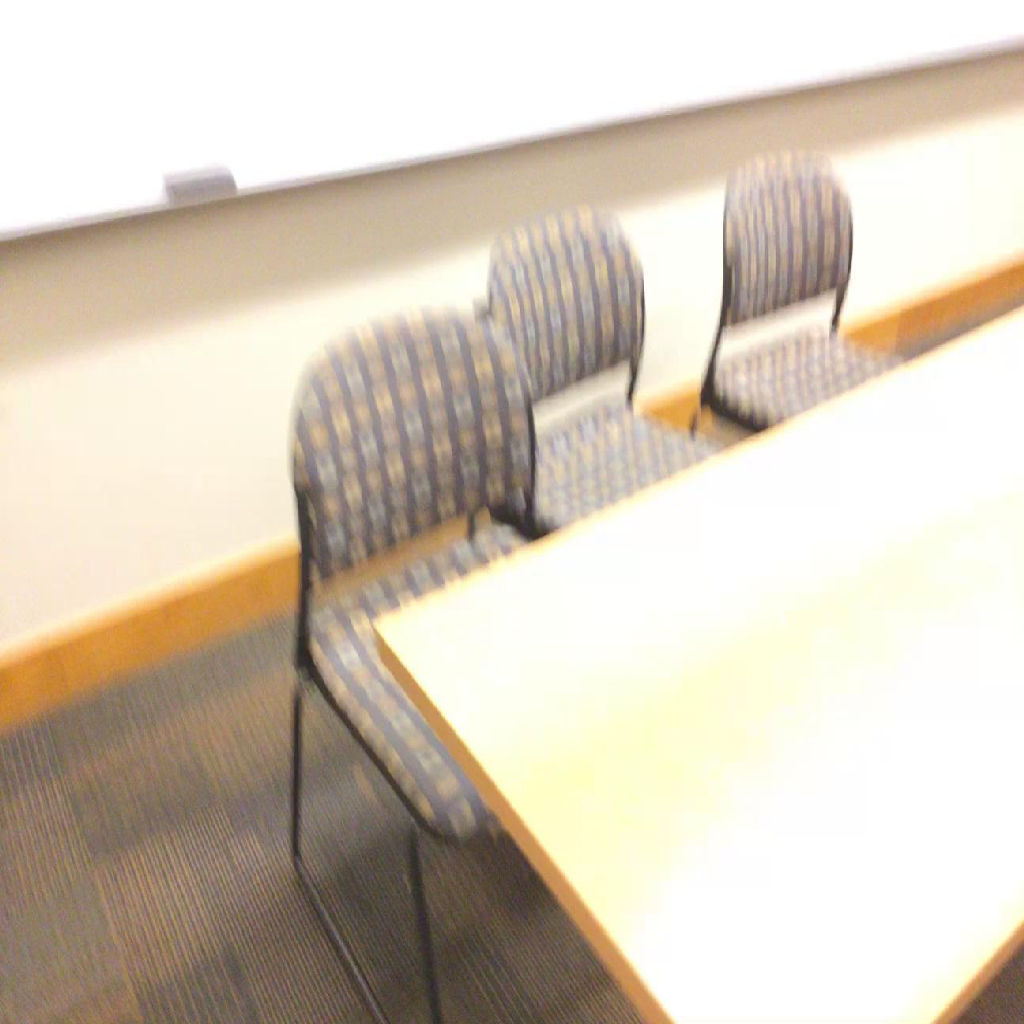}};
\node[image,right=of frame58] (frame59)
    {\includegraphics[width=\linewidth, height=\linewidth]{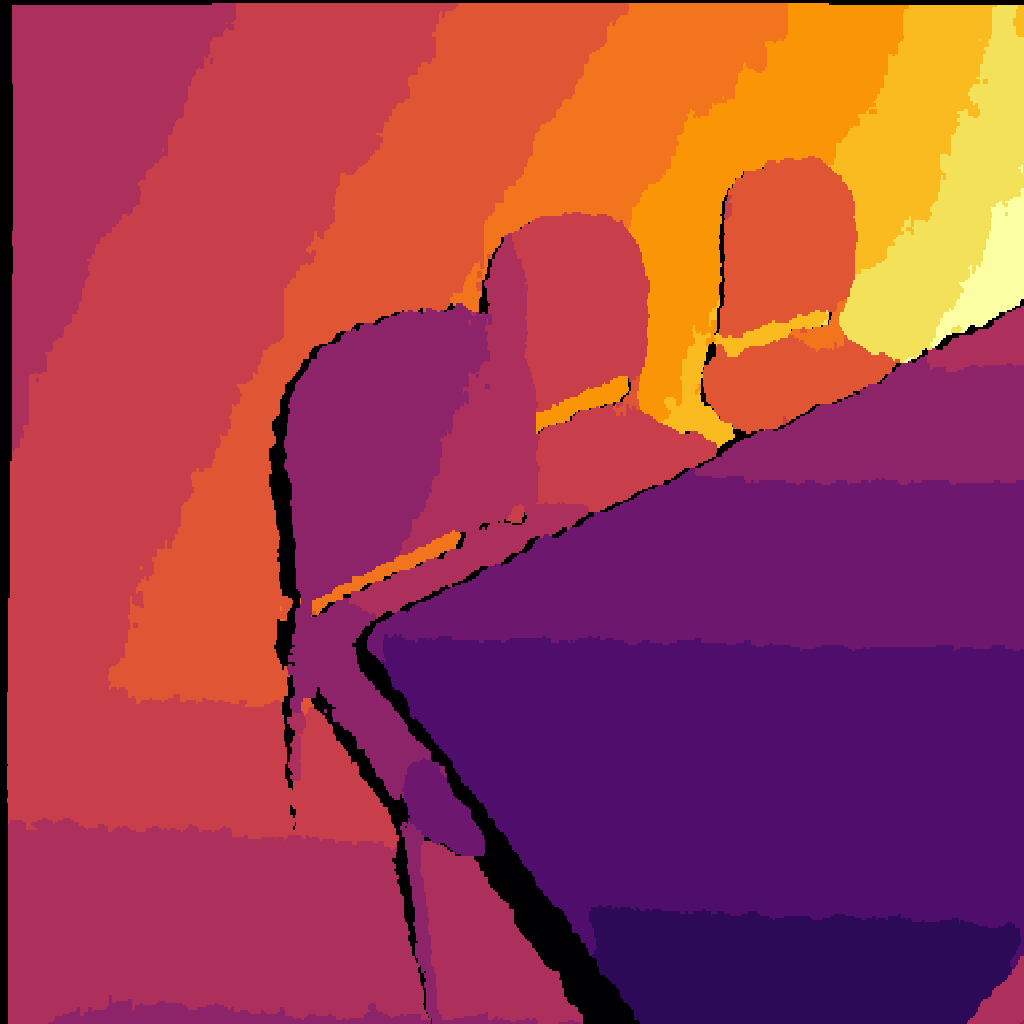}};
\node[image,right=of frame59] (frame60)
    {\includegraphics[width=\linewidth,height=\linewidth]{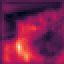}};
\node[image,right=of frame60] (frame61)
    {\includegraphics[width=\linewidth,height=\linewidth]{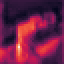}};

\node[image,below=of frame58] (frame62)
    {\includegraphics[width=\linewidth,height=\linewidth]{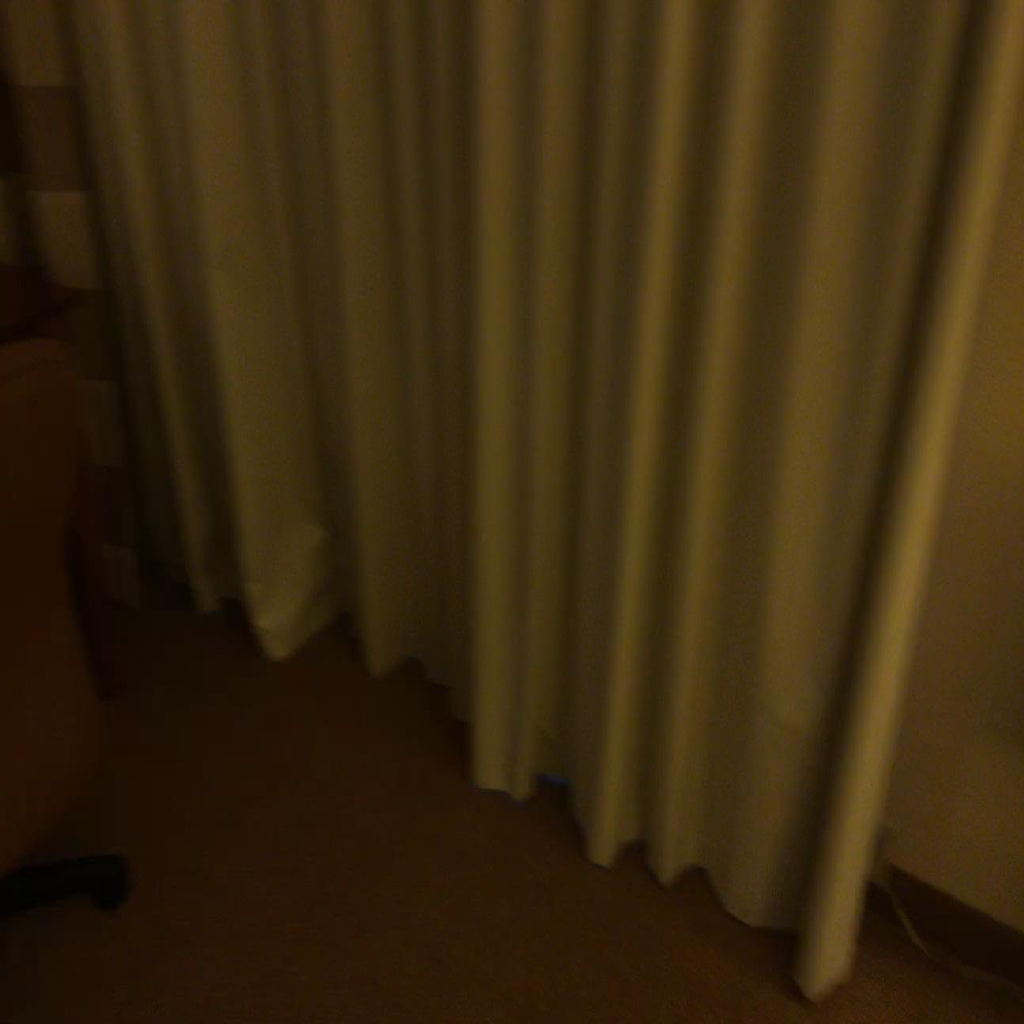}};
\node[image,right=of frame62] (frame63)
    {\includegraphics[width=\linewidth, height=\linewidth]{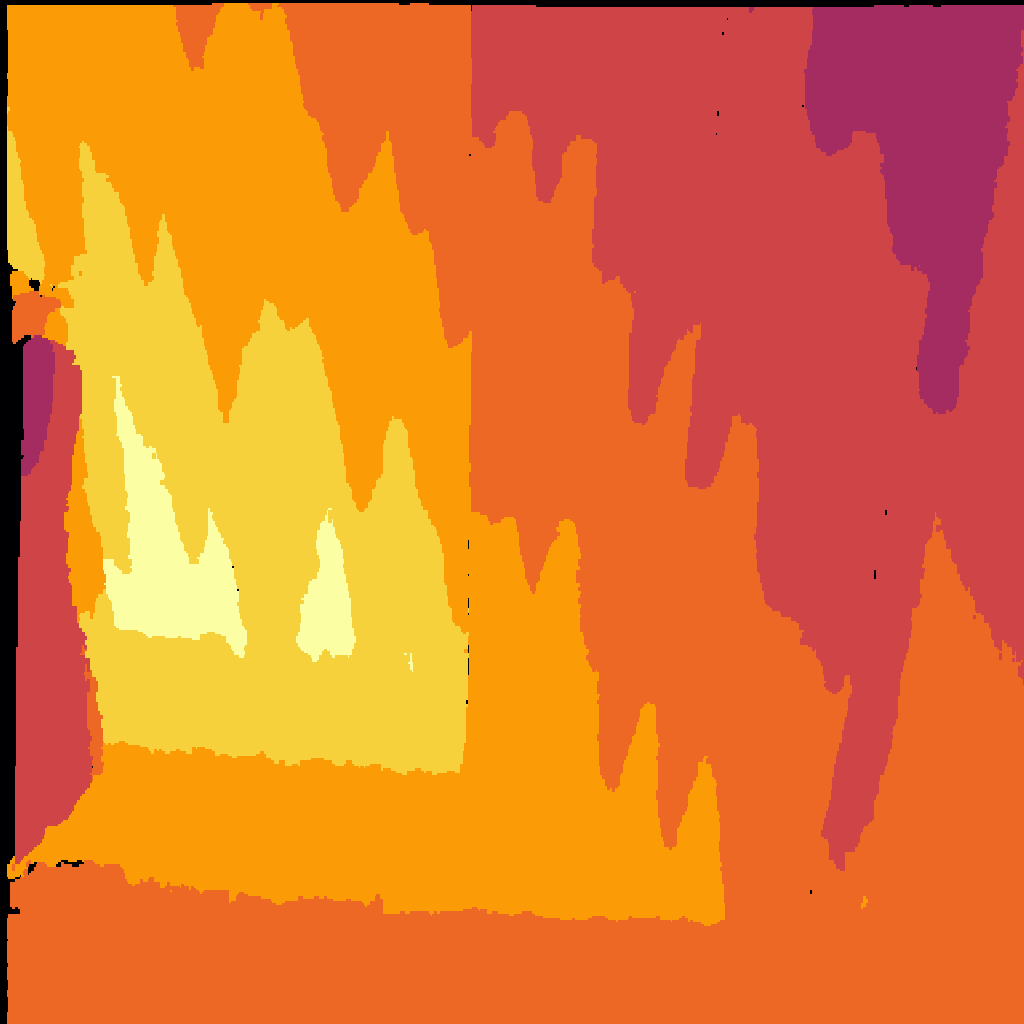}};
\node[image,right=of frame63] (frame64)
    {\includegraphics[width=\linewidth,height=\linewidth]{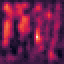}};
\node[image,right=of frame64] (frame65)
    {\includegraphics[width=\linewidth,height=\linewidth]{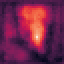}};

\node[image,below=of frame62] (frame66)
    {\includegraphics[width=\linewidth,height=\linewidth]{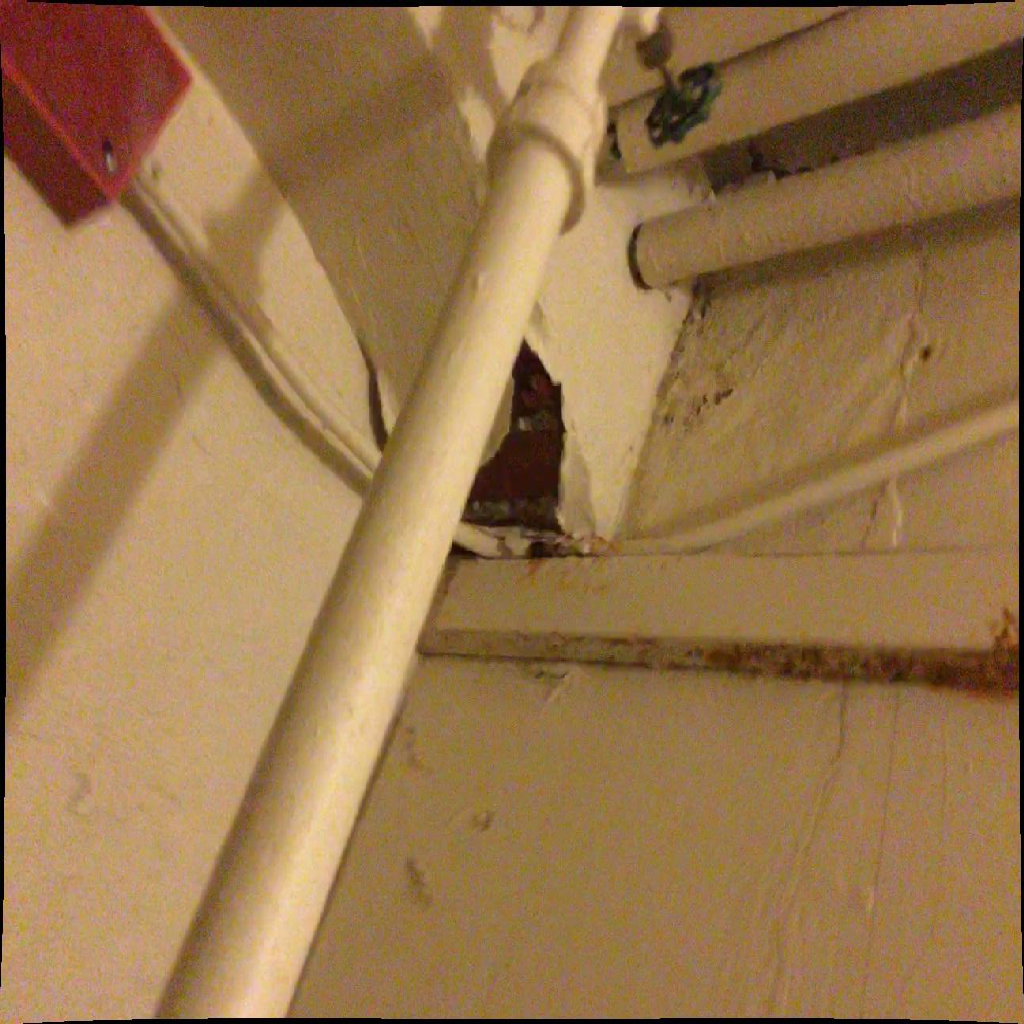}};
\node[image,right=of frame66] (frame67)
    {\includegraphics[width=\linewidth, height=\linewidth]{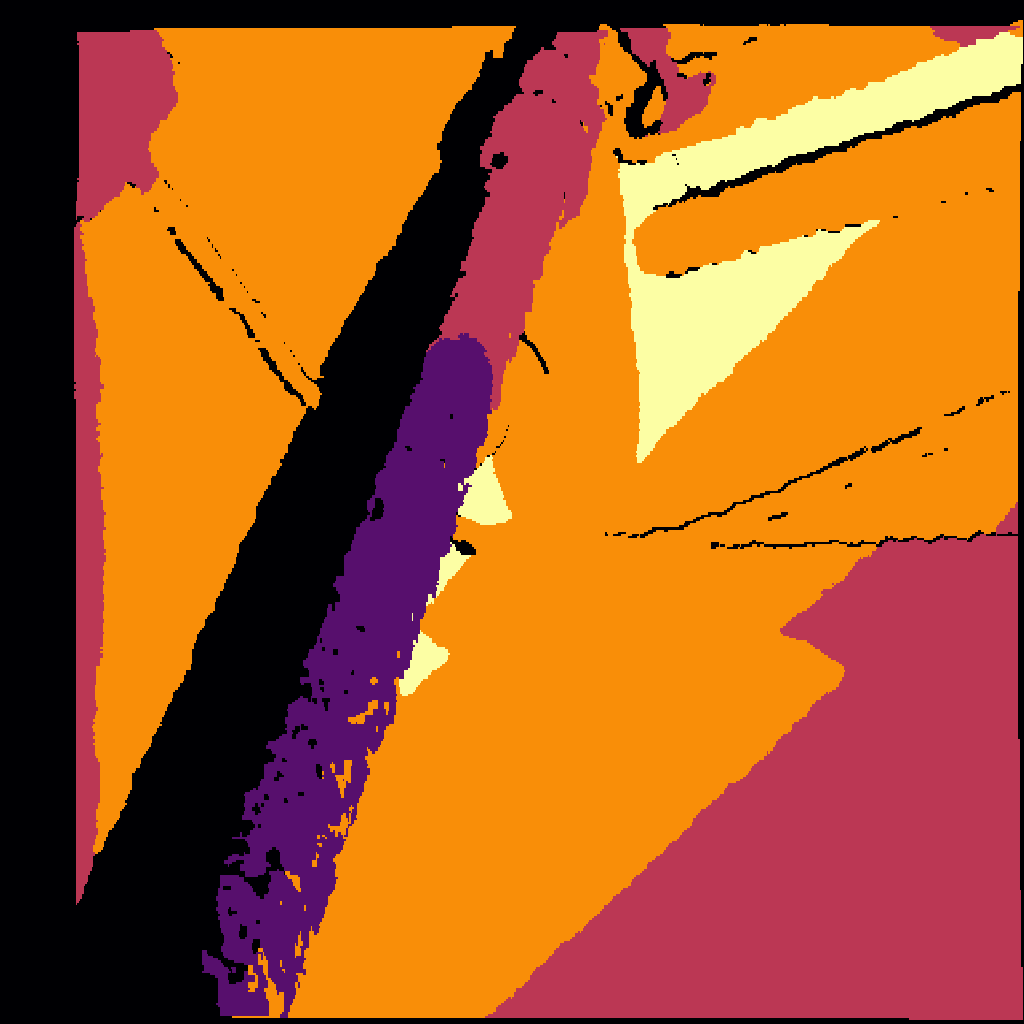}};
\node[image,right=of frame67] (frame68)
    {\includegraphics[width=\linewidth,height=\linewidth]{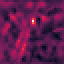}};
\node[image,right=of frame68] (frame69)
    {\includegraphics[width=\linewidth,height=\linewidth]{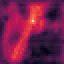}};

\node[image,below=of frame66] (frame70)
    {\includegraphics[width=\linewidth,height=\linewidth]{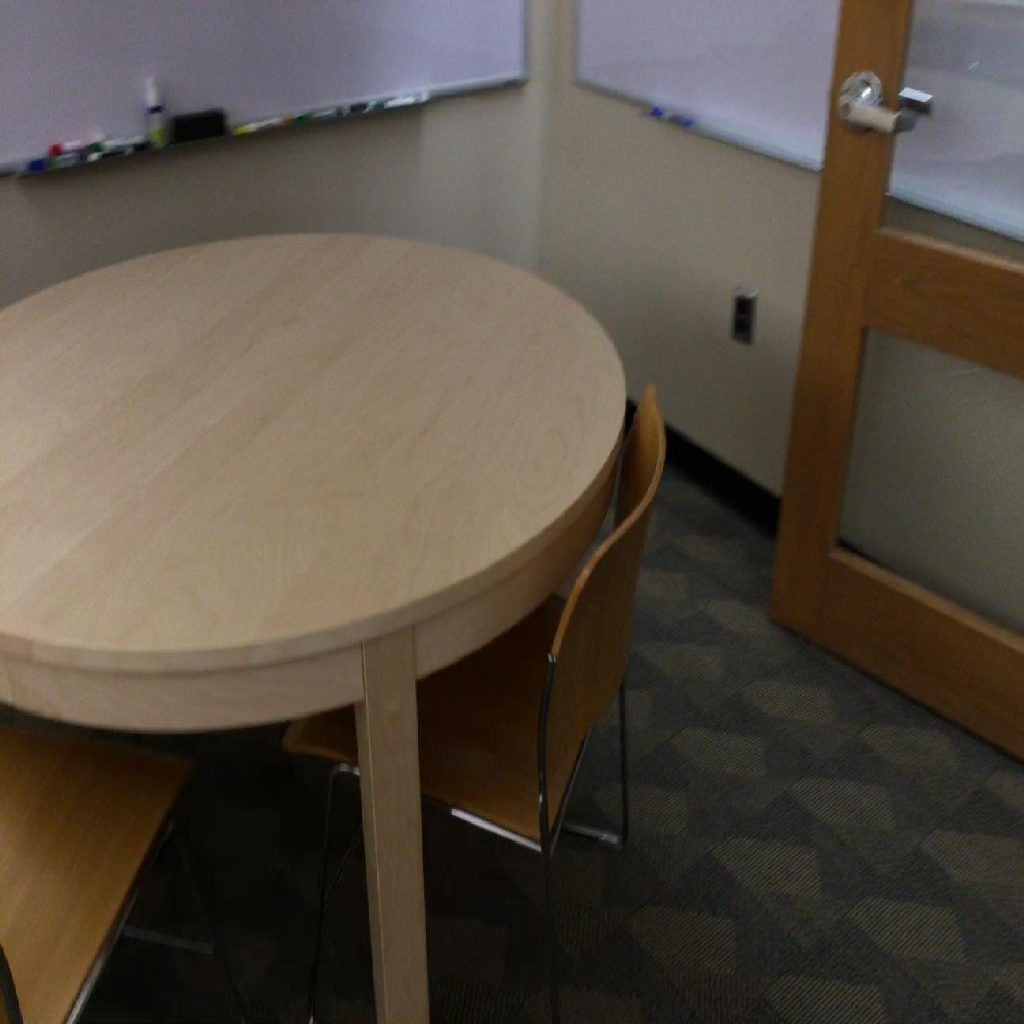}};
\node[image,right=of frame70] (frame71)
    {\includegraphics[width=\linewidth, height=\linewidth]{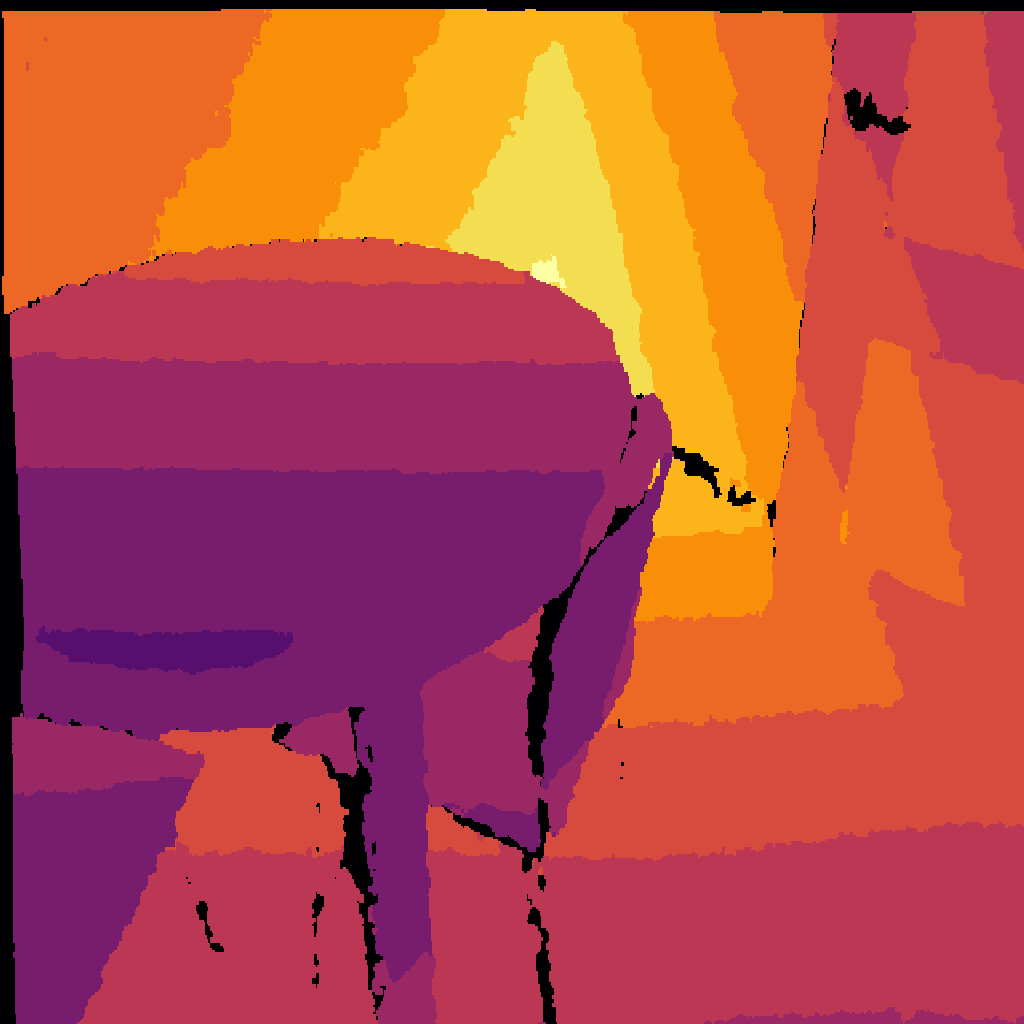}};
\node[image,right=of frame71] (frame72)
    {\includegraphics[width=\linewidth,height=\linewidth]{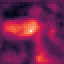}};
\node[image,right=of frame72] (frame73)
    {\includegraphics[width=\linewidth,height=\linewidth]{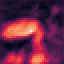}};

\node[image,below=of frame50] (frame74)
    {\includegraphics[width=\linewidth,height=\linewidth]{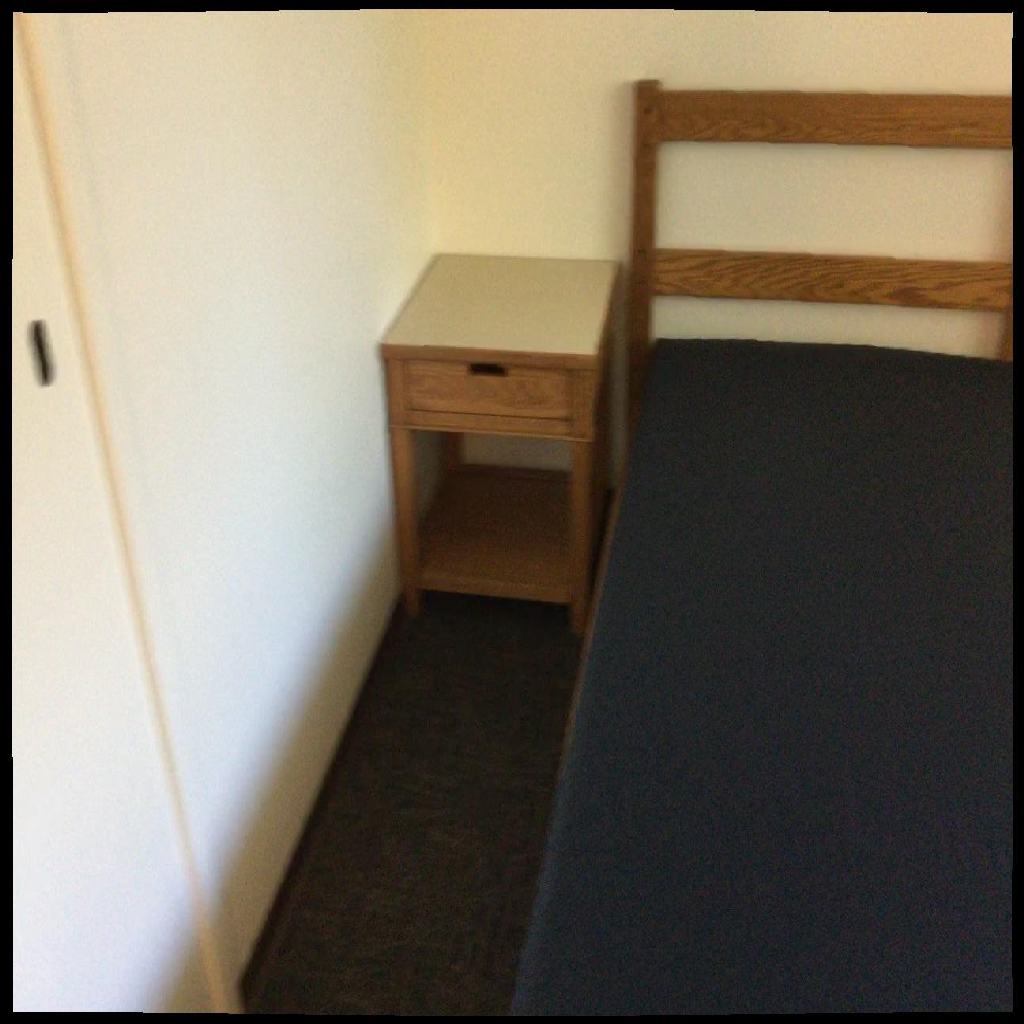}};
\node[image,right=of frame74] (frame75)
    {\includegraphics[width=\linewidth, height=\linewidth]{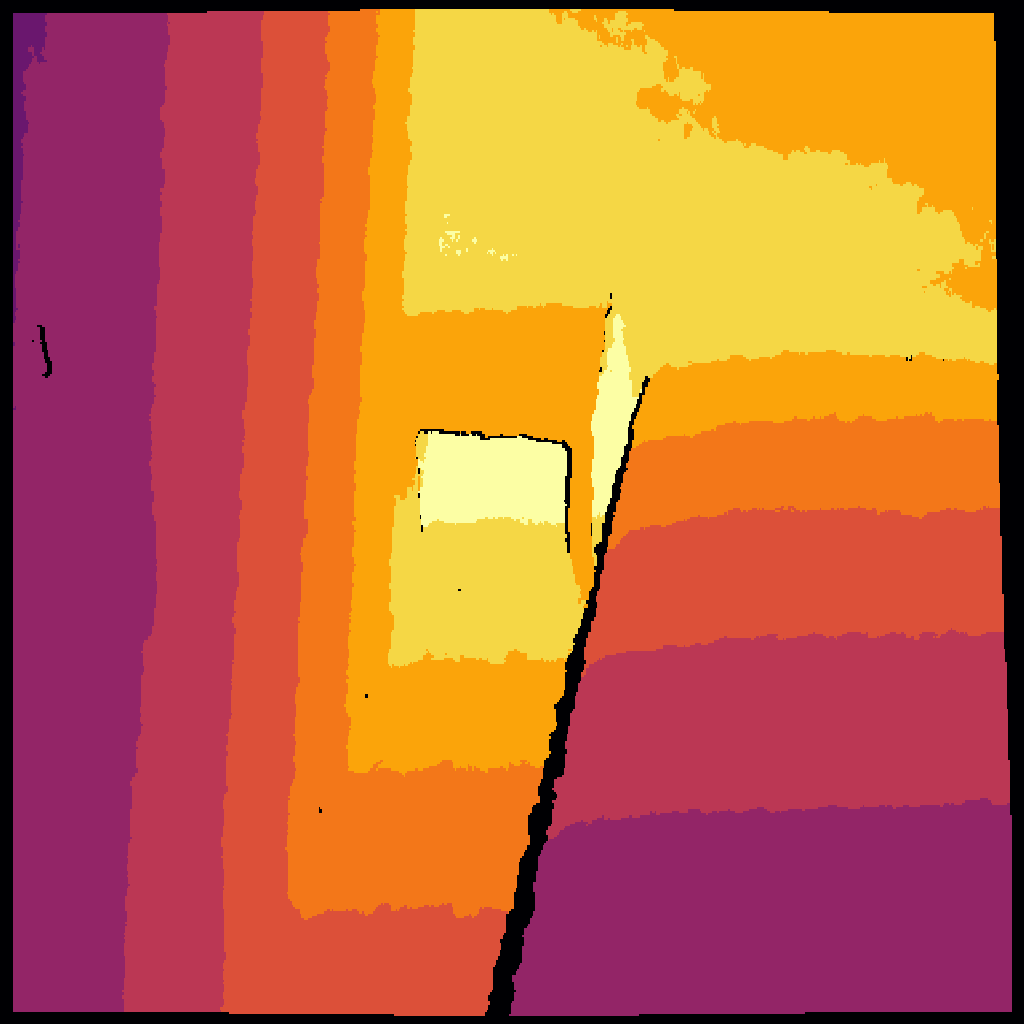}};
\node[image,right=of frame75] (frame76)
    {\includegraphics[width=\linewidth,height=\linewidth]{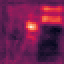}};
\node[image,right=of frame76] (frame77)
    {\includegraphics[width=\linewidth,height=\linewidth]{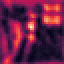}};

\node[image,below=of frame74] (frame78)
    {\includegraphics[width=\linewidth,height=\linewidth]{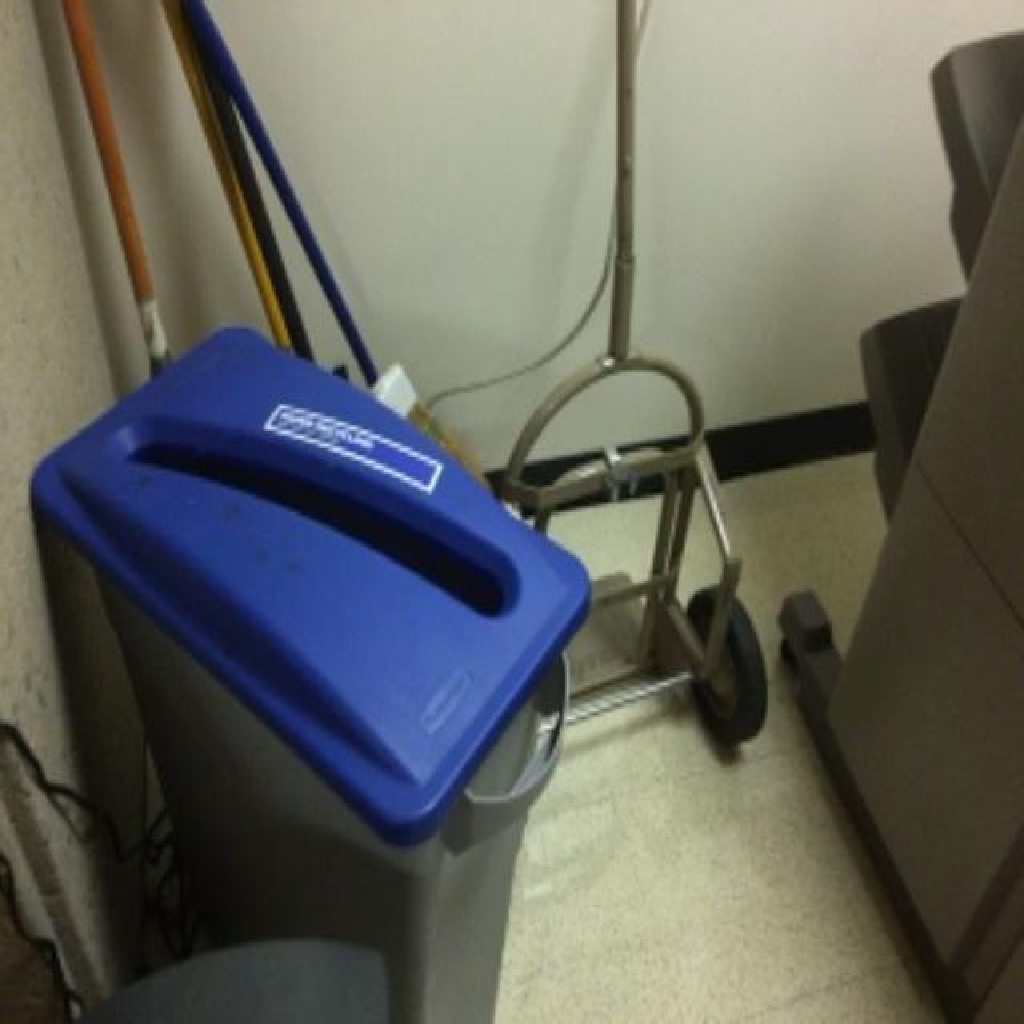}};
\node[image,right=of frame78] (frame79)
    {\includegraphics[width=\linewidth, height=\linewidth]{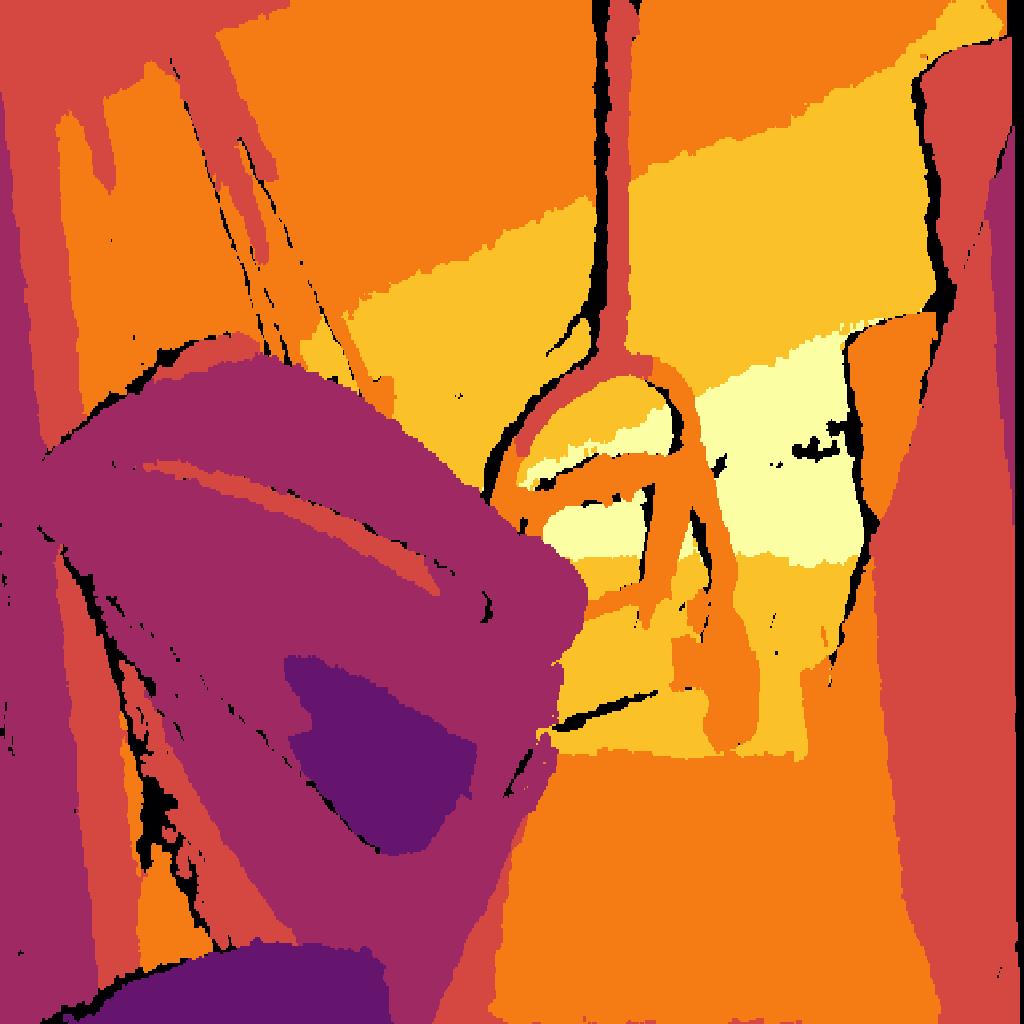}};
\node[image,right=of frame79] (frame80)
    {\includegraphics[width=\linewidth,height=\linewidth]{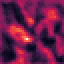}};
\node[image,right=of frame80] (frame81)
    {\includegraphics[width=\linewidth,height=\linewidth]{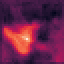}};
    
\node[image,below=of frame78] (frame82)
    {\includegraphics[width=\linewidth,height=\linewidth]{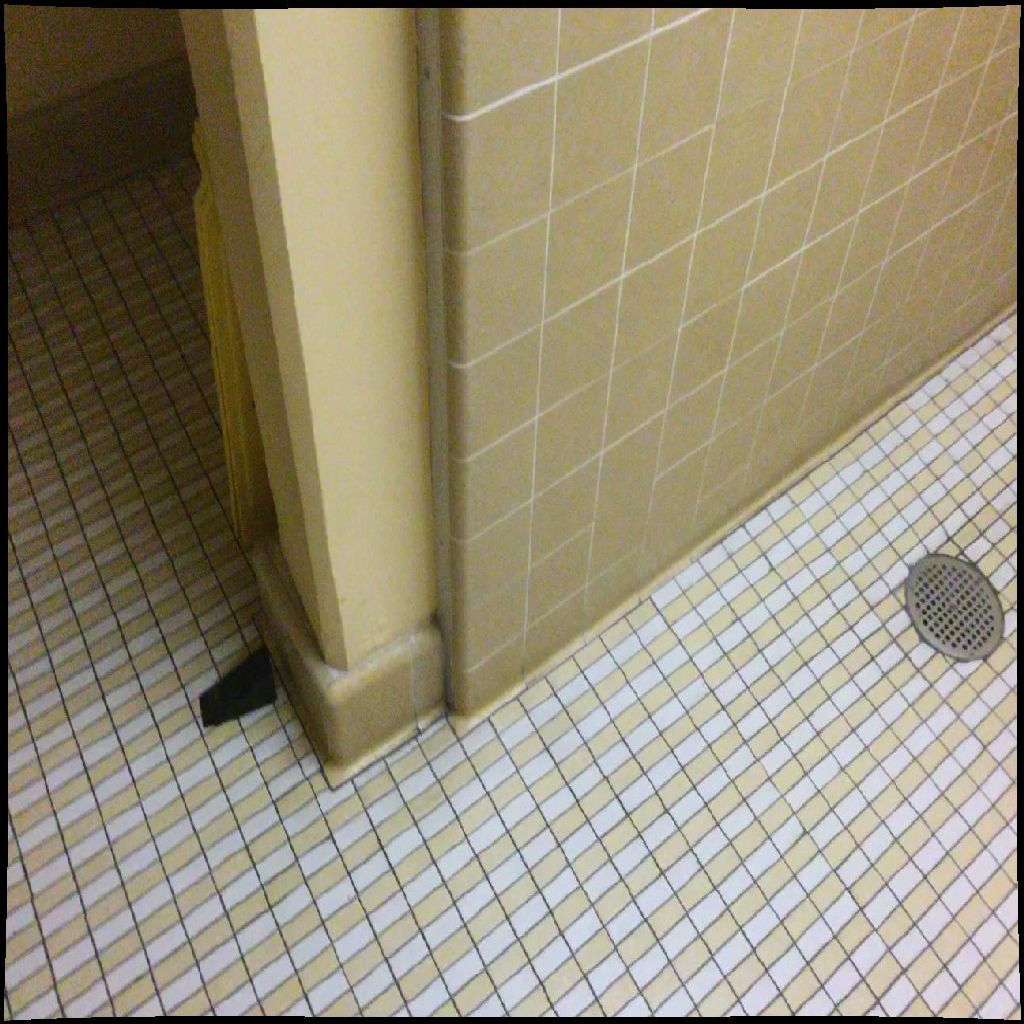}};
\node[image,right=of frame82] (frame83)
    {\includegraphics[width=\linewidth, height=\linewidth]{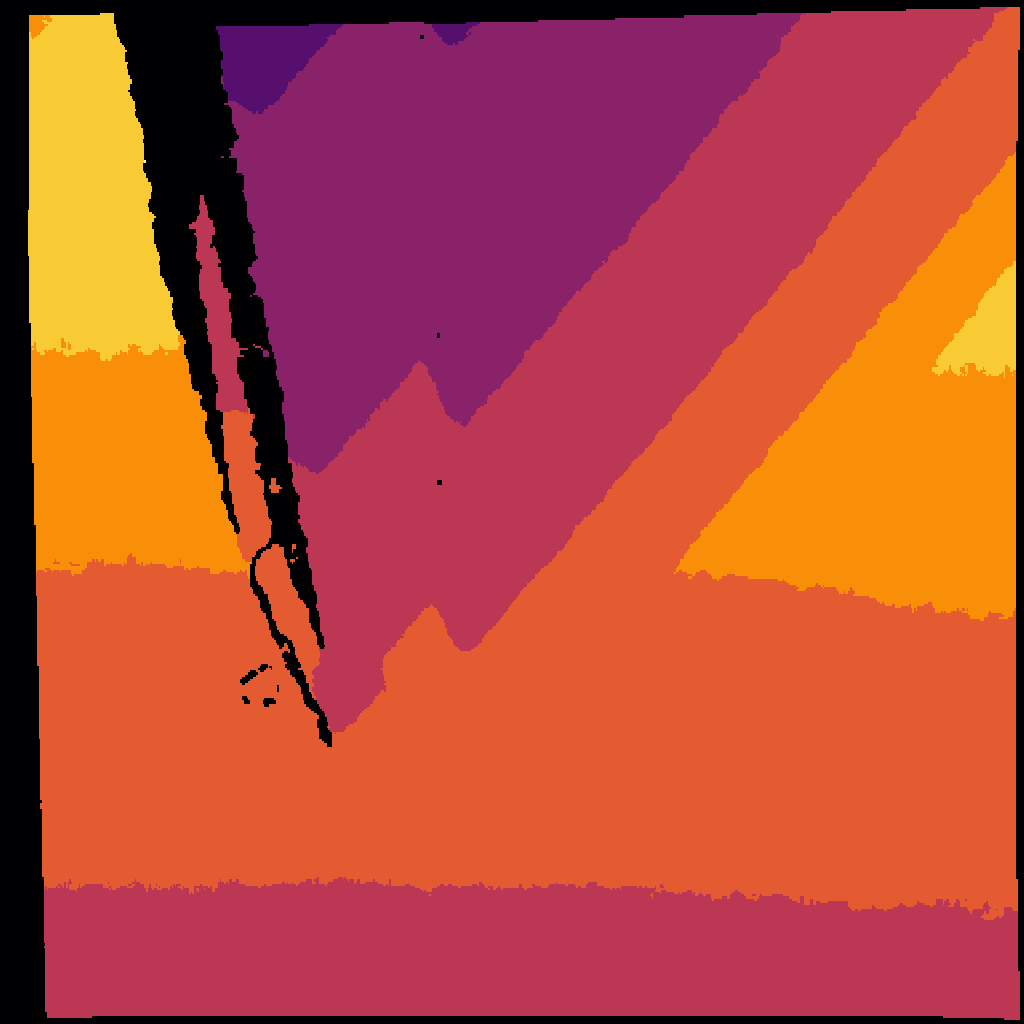}};
\node[image,right=of frame83] (frame84)
    {\includegraphics[width=\linewidth,height=\linewidth]{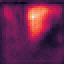}};
\node[image,right=of frame84] (frame85)
    {\includegraphics[width=\linewidth,height=\linewidth]{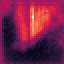}};
    
\node[image,below=of frame82] (frame86)
    {\includegraphics[width=\linewidth,height=\linewidth]{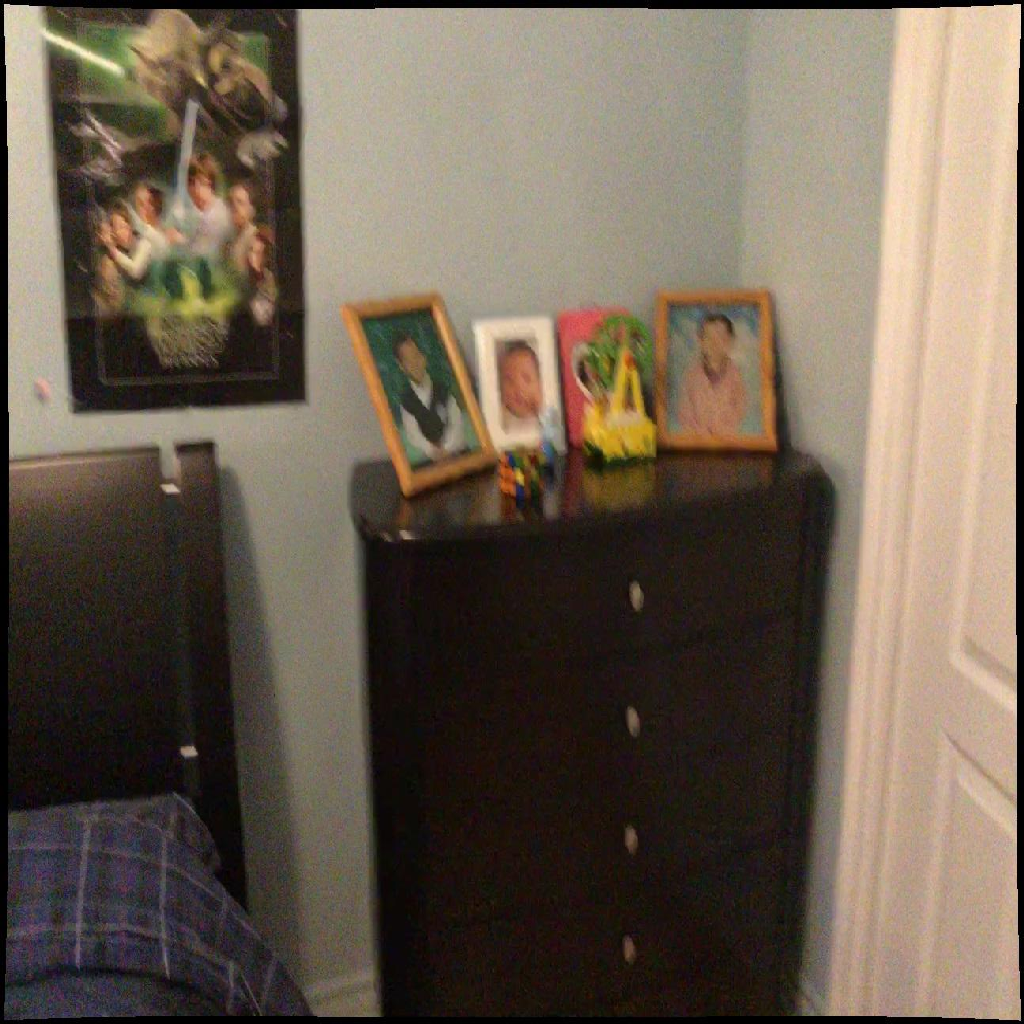}};
\node[image,right=of frame86] (frame87)
    {\includegraphics[width=\linewidth, height=\linewidth]{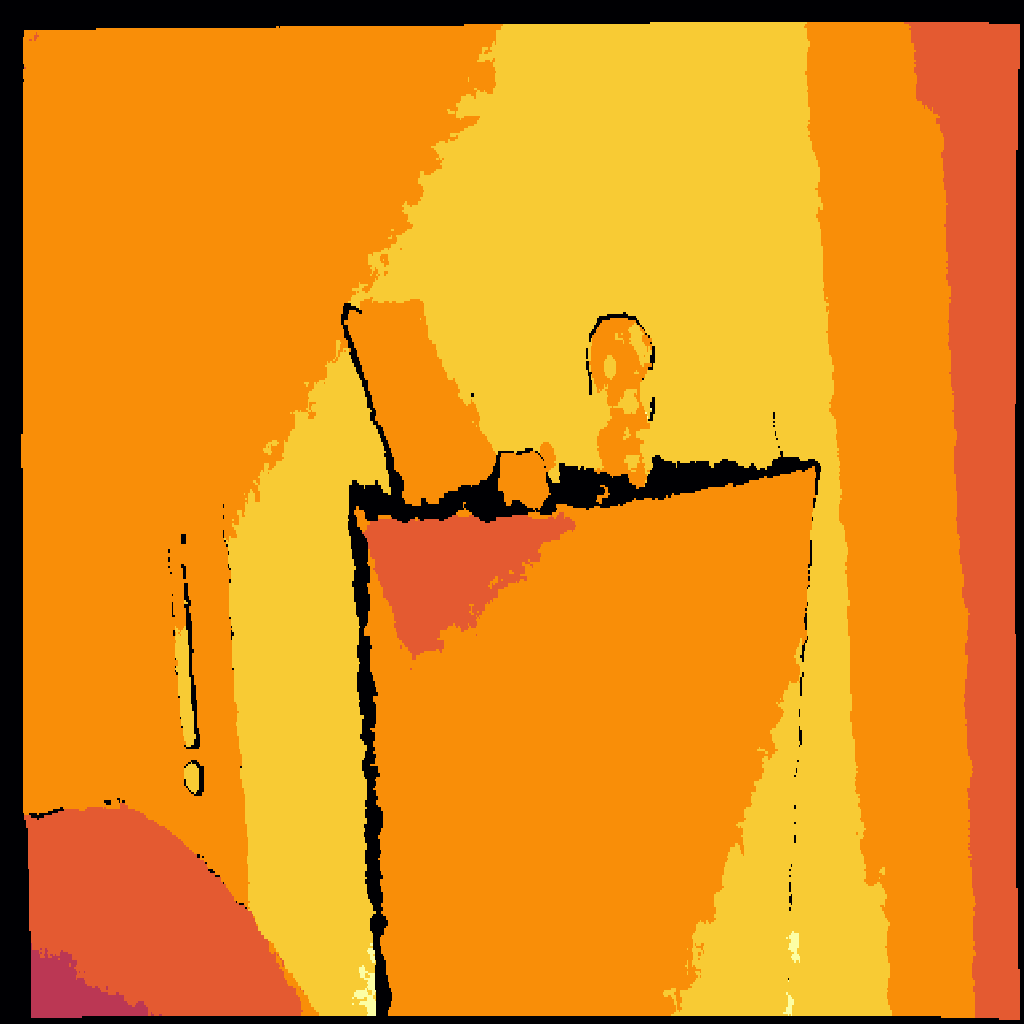}};
\node[image,right=of frame87] (frame88)
    {\includegraphics[width=\linewidth,height=\linewidth]{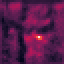}};
\node[image,right=of frame88] (frame89)
    {\includegraphics[width=\linewidth,height=\linewidth]{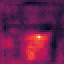}};

\node[text,above= of frame1]  {\scriptsize \vphantom{p}Input Image\vphantom{p}};
\node[text,above= of frame222]  {\scriptsize \vphantom{p}PixPro\vphantom{p}};

\node[text,above= of frame2]  {\scriptsize \vphantom{p}Disparity map\vphantom{p}};
\node[text,above= of frame3]  {\scriptsize \vphantom{p}PixDepth\vphantom{p}};

\node[text,above= of frame14]  {\scriptsize \vphantom{p}Input Image\vphantom{p}};
\node[text,above= of frame15]  {\scriptsize \vphantom{p}Disparity map\vphantom{p}};

\node[text,above= of frame16]  {\scriptsize \vphantom{p}PixPro\vphantom{p}};
\node[text,above= of frame17]  {\scriptsize \vphantom{p}PixDepth\vphantom{p}};

\end{tikzpicture}
}
    \vspace{-1em}
    \caption{
    Comparison of the quality of the learned representations on images from the ScanNet dataset.
    The bright pixels in the similarity maps correspond to the reference feature vector, which is compared to all other vectors forming the features map. Note that these images are part of the validation set (unseen during training), and that the depth map is given for comparison purposes, it is not fed into the encoder.
    }
    \label{fig:results1bis}
\end{figure*}

In this supplementary material we provide more examples of Representation Quality Evaluation experiment described in Section~4 of the main paper. Figure~\ref{fig:results1} shows extra results comparing on several images from ImageNet using networks pre-trained on a subset of the MIDAS training set. The depth maps were computed using MIDAS. Figure~\ref{fig:results1bis} shows the same experiment but on images from ScanNet  while using a network pre-trained on ImageNet. The idea of this variations of training datasets is to use images that are not used during training for evaluation.

